\documentclass[pmlr,twocolumn,10pt]{jmlr}

\usepackage{booktabs}
\usepackage{siunitx}

\newcommand{\equal}[1]{{\hypersetup{linkcolor=black}\thanks{#1}}}

\theorembodyfont{\upshape}
\theoremheaderfont{\scshape}
\theorempostheader{:}
\theoremsep{\newline}

\usepackage{amsmath}
\usepackage{multirow}
\usepackage{graphicx}
\usepackage{comment}
\usepackage{graphicx}
\usepackage{stmaryrd}
\usepackage{dsfont}
\usepackage{pifont}\usepackage{algorithm}
\usepackage{algpseudocode}
\usepackage{cleveref}
\usepackage{colortbl}

\newcommand{\m}{\mathbf}

\usepackage{listings}
\usepackage{enumitem} \usepackage{tablefootnote}

\newcolumntype{I}{>{\itemize}X <{\enditemize}}

\setlist{nosep}

\jmlrvolume{225}
\jmlryear{2023}
\jmlrsubmitted{LEAVE UNSET}
\jmlrpublished{LEAVE UNSET}
\jmlrworkshop{Machine Learning for Health (ML4H) 2023} 

\title[Step-wise Embeddings for Clinical Time-Series]{On the Importance of Step-wise Embeddings for \\ Heterogeneous Clinical Time-Series}

\author{\Name{Rita Kuznetsova$^1$}\equal{These authors contributed equally} \Email{mkuznetsova@ethz.ch}\\
\Name{Alizée Pace$^{1,2,3}$}\footnotemark[1] \Email{alpace@ethz.ch}\\
\Name{Manuel Burger$^1$}\footnotemark[1] \Email{burgerm@ethz.ch} \\
\Name{Hugo Yèche$^1$} \Email{hyeche@ethz.ch} \\
\Name{Gunnar Rätsch$^{1,2,4,5,6}$} \Email{raetsch@inf.ethz.ch} \\
\addr{$^1$Department of Computer Science, ETH Zürich, Switzerland
}\\
\addr{$^2$ETH AI Center, ETH Zürich, Switzerland}\\
\addr{$^3$Max Planck Institute for Intelligent Systems, Tübingen, Germany}\\
\addr{$^4$Medical Informatics Unit, Zürich University Hospital, Zürich, Switzerland}\\
\addr{$^5$Swiss Institute of Bioinformatics, Zurich, Switzerland}\\ 
\addr{$^6$Department of Biology, ETH Zürich, Zürich, Switzerland}\\
}

\begin{document}

\maketitle

\begin{abstract}
Recent advances in deep learning architectures for sequence modeling have not fully transferred to tasks handling time-series from electronic health records. In particular, in problems related to the Intensive Care Unit (ICU), the state-of-the-art remains to tackle sequence classification in a tabular manner with tree-based methods. Recent findings in deep learning for tabular data are now surpassing these classical methods by better handling the severe heterogeneity of data input features. Given the similar level of feature heterogeneity exhibited by ICU time-series and motivated by these findings, we explore these novel methods' impact on clinical sequence modeling tasks. By jointly using such advances in deep learning for tabular data, our primary objective is to underscore the importance of step-wise embeddings in time-series modeling, which remain unexplored in machine learning methods for clinical data. On a variety of clinically relevant tasks from two large-scale ICU datasets, MIMIC-III and HiRID, our work provides an exhaustive analysis of state-of-the-art methods for tabular time-series as time-step embedding models, showing overall performance improvement. In particular, we evidence the importance of feature grouping in clinical time-series, with significant performance gains when considering features within predefined semantic groups in the step-wise embedding module.
\end{abstract}
\begin{keywords}
Deep Learning, Healthcare, Time-Series, Step-wise Embeddings, Feature Groups.
\end{keywords}

\section{Introduction}
Recent years have seen the development of deep learning architectures for Electronic Health Records (EHRs), which explore machine learning solutions for various clinical prediction tasks such as organ failure prediction \citep{hyland2020, tomavsev2019clinically}, treatment effect estimation \citep{bica2020estimating} or prognostic modeling \citep{choi2016doctor}. Most work in this area primarily focuses on either modifying the backbone sequence model~\citep{DBLP:conf/icml/HornMBRB20, xu2018raim} or investigating modifications to the training objective~\citep{yeche2021neighborhood, yeche2022temporal, cheng2023timemae}. Still, the performance gap between proposed deep learning methods and tree-based approaches remains significant~\citep{yeche2021, hyland2020}. 

Recent work for early prediction of acute kidney injury using sparse multivariate time-series \citep{tomavsev2021use} shows that enhancing the time-step embedding neural network architectures, i.e simple replacement of linear layer to neural network for the input feature space preprocessing, yields significant performance gain. Concurrently, the state-of-the-art on tabular data, which relied on boosted tree methods~\citep{ke2017lightgbm, chen2016xgboost, freund1999short}, has been surpassed by recent development in the field of deep learning~\citep{gorishniy2021revisiting, gorishniy2022embeddings}.  
Despite these observations, recent research in EHRs methods predominantly showcases the development of more powerful backbone sequence models, rather than investigating the influence of step-wise embedding modules. If some approaches have used feature embeddings, with their primary focus being on evaluating the effect of self-supervised pre-training~\citep{tipirneni2022self}, a comprehensive evaluation of how feature embedding efforts influence downstream performance is yet to be extensively studied.

Motivated by these observations, our main objective is to showcase the \textbf{significance of embedding architectures in clinical time-series analysis}. To achieve this, we conduct an extensive evaluation and comparison of various embedding architectures specifically designed for tabular data, with a focus not on optimizing the backbone sequence model, but rather on optimizing the step-wise embedding module. We find that we obtain timestep representations that serve as an expressive input to downstream sequence models -- which boosts the overall performance of deep learning methods on clinical time-series data. Our work is thus orthogonal and complementary to the design of backbone architectures~\citep{DBLP:conf/icml/HornMBRB20} or of loss functions for supervised~\citep{yeche2022temporal} and unsupervised learning~\citep{yeche2021neighborhood}.

\begin{figure}[t]
    \centering
    \includegraphics[width=\linewidth]{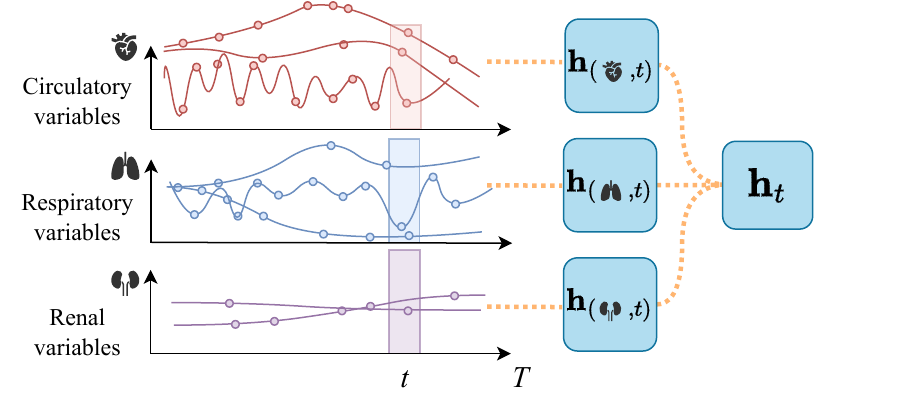}
    \caption{\textbf{Schematic time-step embedding architecture}: features interact within predefined semantic groups before being aggregated into time-step embeddings.}
    \label{fig:overview}
    \vspace{-0.5cm}
\end{figure}

Second, our study demonstrates the importance of \textbf{feature groupings}~\citep{imrie2022, masoomi2020} in clinical time-series. In the medical field, it is common to not consider measurement interactions individually but through predefined semantic groups of features~\citep{kelly2009multiple, meira2001cancer}. EHR data consist of multivariate time-series exhibiting such heterogeneity, with variables collected from different data sources and relating to different organ systems. These structures, determined by prior clinical knowledge, delineate feature groups tied to {medical concepts} or {modalities}, such as measurement types or organs, which we incorporate into embedding modules. Results demonstrate considering features \textit{in the context of their semantic modality} to improve performance. We illustrate the optimal embedding pipeline uncovered by our work in \figureref{fig:overview}: features interact within groups before being aggregated into time-step embeddings and input to a sequential deep learning module for end-to-end training. This scheme additionally enables the interpretability of model results at a semantic group level. Thus, we also explore how disentangling medical concepts could enhance the interpretability of the model’s decision-making.

\paragraph{Contributions} The main contributions of this paper are the following: (1) First, we provide an extensive benchmark of embedding architectures for clinical prediction tasks. To the best of our knowledge, no prior work has considered applying the developments from the tabular deep learning literature to the heterogeneous time-series nature of clinical data. (2) Our exhaustive analysis allows us to draw important conclusions that semantically grouping features, especially related to organ systems, greatly enhance prediction performance. (3) Finally, combining these insights, our systematic study sets a new state-of-the-art performance on different established clinical tasks and datasets.

\section{Related work}\label{sec:rel_work}

\paragraph{Time-series feature embedding} Despite developments in model architectures for supervised clinical time-series tasks~\citep{DBLP:conf/icml/HornMBRB20, zhang2021graph}, deep learning methods still show performance limitations on the highly heterogeneous, sparse time-series nature of intensive care unit data~\citep{yeche2021, hyland2020}. Recent work has, however, demonstrated promising improvements with the introduction of feature embedding layers before the sequence model, together with auxiliary objectives~\citep{tomavsev2021use, tomavsev2019clinically}. 
This research mirrors recent progress in the field of deep learning for tabular data~\citep{gorishniy2021revisiting, gorishniy2022embeddings}, which significantly outperforms state-of-the-art methods by combining transformer-based approaches with embeddings of tabular data rows. 
We note that a separate line of research explores self-supervised pre-training methodologies for both clinical time-series representation learning \citep{tipirneni2022self,labach2023effective}and tabular deep learning \citep{yin2020tabert,huang2020tabtransformer,kossen2021self,somepalli2022saint}. While we focus on end-to-end supervised training in the present benchmark, we note that this constitutes a promising avenue for further work in clinical time-series feature embedding.
\paragraph{Feature groups within embeddings} 
Recent work on tabular data embeddings highlight the importance of handling features of categorical or numerical types through distinct architectures \citep{huang2020tabtransformer,arik2021tabnet}. This motivates our benchmarking study on incorporating additional feature structures, such as measurement or organ type, within the embedding layers. Most research on EHR data modeling focuses on extracting temporal trends~\citep{luo2016predicting, ding2021unsupervised} for patient phenotyping~\citep{aguiar2022learning,qin2023t} from entire time-series. To the best of our knowledge, this work is the first attempt to consider and demonstrate the impact of global feature groupings at a time-step level on prediction performance. 

We refer the reader to \appendixref{appendix: prior_work} for further discussion of related work.

\section{Method}\label{sec:methods}
\begin{figure*}[t]
  \centering
  \includegraphics[width=\linewidth]{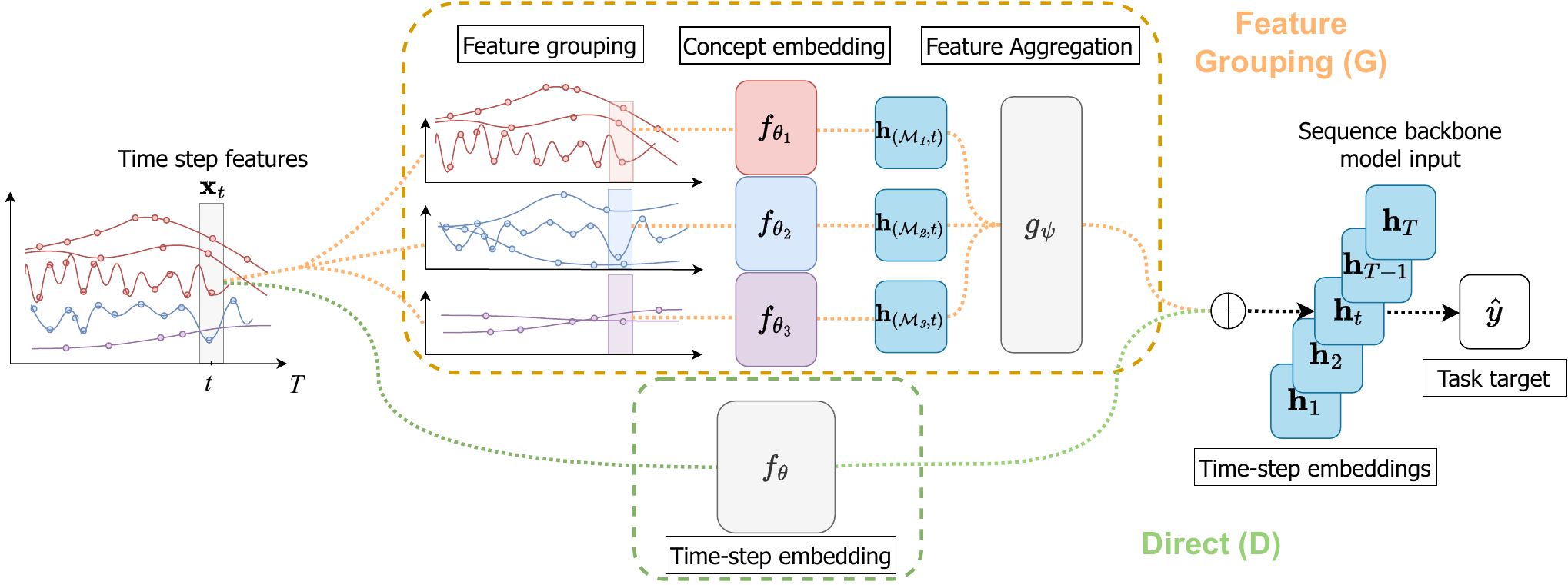}
  \caption{\textbf{Pipeline overview.} The entire set of features $\m{x}_t$ for time-step $t$ 
    is: (D) preprocessed to directly form a time-step embedding $\m{h}_t$ (green line); (G) grouped to form concept embeddings $\m{h}_{(\mathcal{M}_k, t)}$, which are aggregated to create a final time-step embedding $\m{h}_t$ (yellow line). The resulting time-step embeddings are then passed to the backbone model. The whole pipeline is trained in an end-to-end fashion to predict the task target $\hat{y}$.}
  \label{fig:pipeline-detailed}
\end{figure*}

We summarize the overall deep learning pipeline benchmarked in this work in \figureref{fig:pipeline-detailed}, followed by an in-depth explanation of each component in this section.

\subsection{Notation}\label{sec:notation} 

We define a patient stay in the intensive care unit as a multivariate time-series $\m{X} = [\m{x}_{1}, \ldots, \m{x}_{T}]$, where $T$ is the length of a given stay. Each time-step is $\m{x}_t = [x_{{(1,t)}}, \ldots, x_{{(d,t)}}] \in \mathbb{R}^d$.
Depending on the specific task, for each patient stay $\m{X}$ we have either an associated label vector $\m{y} \in \mathbb{R}^T$ (per each time-step) or a single label $\m{y} \in \mathbb{R}$ that corresponds to the entire patient stay -- see \sectionref{sec:exp_setup} for an overview of studied tasks and datasets.

We consider step-wise embedding architectures under two scenarios: first, as a function applied to the entire feature space $\m{x}_t$, referred to as \emph{direct} (D); second, we propose to apply them separately to distinct feature groups, referred to as \emph{feature grouping} (G). In the latter case, there are several ways to group the $d$ observed variables for each time-step $\m{x}_t $ based on their assignment to a particular medical concept from a set of $K$ concepts $\{\mathcal{M}_1, \ldots, \mathcal{M}_K\}$, such that all variables are assigned to a single concept: $\bigcup_{k=1}^K \mathcal{M}_k = \{1, \ldots, d\}$ and $\forall k \neq k^{'}: \mathcal{M}_{k^{'}}\cap\mathcal{M}_k = \emptyset$. In the context of ICU-related tasks, we define the splitting of features into concept groups by leveraging the prior knowledge: \emph{organ}, \emph{measurement type} (laboratory values, observations, treatments, etc.) and \emph{variable type} as shown by \citet{tomavsev2019clinically}. The exact groups are provided in~\appendixref{appendix:splittings}. We group the features on a time-step level $t$, and we denote the subset of features belonging to the concept $\mathcal{M}_k$ as $\m{x}_{(\mathcal{M}_k, t)}$. For each $k$, we learn a representation of $\m{h}_{(\mathcal{M}_k, t)}$ that we refer to as \emph{concept embedding}.

\begin{definition}\label{def:concept_emb} Let $f_{\theta_k}$ denote the embedding model for concept $\mathcal{M}_k$, parameterized by $\theta_k$, taking as input the subset of features $\m{x}_{(\mathcal{M}_k, t)}$ and with output $\m{h}_{(\mathcal{M}_k, t)} = f_{\theta_k}\bigl(\m{x}_{(\mathcal{M}_k, t)} \bigr)$. We term the latent representation $\m{h}_{(\mathcal{M}_k, t)}$ as the \emph{concept embedding}.
\end{definition}

\begin{definition}\label{def:ts_emb}
The \emph{time-step embedding} $\m{h}_t$ is a representation of all input features $\m{x}_t$ at a given time-step. This embedding can be obtained through two approaches, as illustrated in \figureref{fig:pipeline-detailed}:
\begin{itemize}
\item[(D)] In the first (direct) scenario, $\m{h}_t = f_{\theta}(\m{x}_t)$, where $f_{\theta}$ is an embedding model parameterized by $\theta$, processing the entire set of features for each time-step $t$.
\item[(G)] In the second (feature grouping) scenario, $\m{h}_t = g_{\psi} \bigl[ \bigl\{\m{h}_{(\mathcal{M}_k, t)}\bigr \}_{k=1}^K \bigr]$, where $g_{\psi}$ is an aggregation function applied to the $K$ concept embeddings of each feature group.
\end{itemize}
\end{definition}

The resulting time-step embedding $\m{h}_t$ is subsequently passed as input to the sequential backbone. In the following, we discuss design choices for feature encoder architectures ($f_\theta$ and $f_{\theta_k}$) and for the aggregation function $g_{\psi}$.

\subsection{Direct time-step embedding}\label{sec:mod_emb}
As first candidates, following~\citep{gorishniy2021revisiting, grinsztajn2022tree}, we consider MLP and ResNet architectures as feature encoders. These are well-studied deep learning models, whose impact on step-wise feature preprocessing remains unexplored in the context of clinical sequence modeling.

We also consider a more advanced architecture borrowed from deep learning for tabular data, the Feature Tokenizer Transformer (FTT~\citep{gorishniy2021revisiting}). This complex encoder consists of two distinct modules. First, the \emph{Feature Tokenizer} (FT) embeds individual features $x_{{(j,t)}}\in \mathbb{R}$ in timestep vector $\m{x}_{t}$ to high-dimensional continuous variables $\m{e}_{(j,t)} \in \mathbb{R}^m$. This module is linear, parametrized  by $\m{W} \in \mathbb{R}^{d \times m}$, such that $\m{e}_{(j,t)} =  \m{x}_t^{T} \m{W}_j$. The final output of the FT module is a matrix $\m{e}_t = \texttt{stack}[\m{e}_{1,t},\ldots,\m{e}_{d,t}] \in \mathbb{R}^{d \times m} $.  Next, the \emph{Transformer} (T) module learns a unique time-step embedding $\m{h}_t$ from matrix $\m{e}_t$, by applying a transformer~\citep{vaswani2017attention} along the $d$ dimension. More specifically, to obtain a global representation, $\m{h}_t$ is the output from a ``classification token'' [\texttt{CLS}]~\citep{devlin2018bert} which is concatenated to the input $\m{e}_t$.

We do not consider unsupervised methods such as factor analysis, standard auto-encoders, and variational auto-encoders for the embedding module design, given reports of them not demonstrating significant performance benefits for ICU data feature embeddings~\citep{tomavsev2019clinically}. Compared to MLP and ResNet, which consider features equally, FTT, through this two-stage modeling, should handle feature heterogeneity more efficiently, a crucial consideration in the context of ICU data.

\subsection{Feature aggregation}\label{sec:aggregation}
As introduced in \sectionref{sec:notation}, in scenario (G), our aim is to explore the impact of embedding distinct groups of features independently. There, we simply use the same architecture for $K$ concept embedding models, each with its own set of parameters $\theta_k$ as in \definitionref{def:concept_emb}.

In terms of aggregation function $g_{\psi}$, designed to combine concepts $\m{h}_{(\mathcal{M}_k,t)}$ into an overall timestep embedding $\m{h}_t$, we consider the choices: mean (or sum) pooling, concatenation \footnote{The concatenation function is not, by itself, an aggregation function. It also presents scalability issues and lacks permutation invariance. Nevertheless, we include it in our study for experimental purposes.}, and attention-based pooling. The latter option additionally offers interpretability of concept-level interactions through attention weight analysis, as discussed in \sectionref{sec:interpretability}.

\subsection{Training}\label{sec:training} 
The entire set of features $\m{x}_t$ for time-step $t$ is preprocessed as shown in \figureref{fig:pipeline-detailed}. The resulting time-step embeddings for each for each $t$ are subsequently fed into the sequential backbone model, which is trained in a supervised manner for the final task's target prediction $\hat{y}$. Consistent with previous approaches\citep{tomavsev2019clinically, tomavsev2021use, gorishniy2021revisiting}, no specific loss for the embeddings was factored in. The primary objective of this study is to demonstrate that a simple step-wise module integrated in standard end-to-end supervised training pipeline can produce significant performance improvements.

\begin{table*}[tb]

\floatconts
  {tab:performance}{
      \caption{\textbf{Performance benchmark for different embedding architectures,} measured through the Area under the Precision-Recall Curve (AUPRC) or Mean Absolute Error (MAE) in hours. Mean and standard deviation are reported over five training runs. Best and overlapping results are highlighted in bold. \emph{Reference Benchmark} results are best as reported by~\citet{yeche2021} and \citet{harutyunyan2019multitask} (We train LightGBM, Transformer and Temporal Convolutional Network (TCN) on MIMIC-III for comparison). \emph{Step-wise encoders} are based on prior work (linear~\cite{yeche2021}, MLP~\cite{tomavsev2019clinically}, ResNet~\cite{tomavsev2019clinically}, and FTT~\cite{gorishniy2021revisiting}) and our proposed concept groups. The \emph{backbone} baseline considers the raw input feature vector at each time-step without any embedding layer.}
    }{
        \footnotesize
        \setlength{\tabcolsep}{3pt}
        \begin{tabular}{l c c c c c c c c}
        \toprule
        Dataset & \multicolumn{5}{c} {HiRID} & \multicolumn{3}{c}{MIMIC-III} \\
        \cmidrule(lr){2-6} \cmidrule(lr){7-9}
        Clinical pred. task & Circ. Fail. & Resp. Fail. & Mort. & LoS & Pheno. & Decomp. & Mort. & LoS \\
        Metric & \textit{AuPRC} $\uparrow$ & \textit{AuPRC} $\uparrow$ & \textit{AuPRC} $\uparrow$ & \textit{MAE} $\downarrow$ & \textit{Bal. Acc.} $\uparrow$ & \textit{AuPRC} $\uparrow$ & \textit{AuPRC} $\uparrow$ & \textit{MAE} $\downarrow$  \\ \midrule
        
\multicolumn{6}{l}{\textbf{Reference Benchmarks}} \\
        \arrayrulecolor{lightgray}\midrule\arrayrulecolor{black}
        LSTM & 32.6 $\pm$ 0.8 & 56.9 $\pm$ 0.3 &  60.0 $\pm$ 0.9 &  60.7 $\pm$ 1.6 & 39.5 $\pm$ 1.2 & 34.4 $\pm$ 0.1 & 48.5 $\pm$ 0.3 & 123.1 $\pm$ 0.2 \\
        Transformer &  35.2 $\pm$ 0.6 & 59.4 $\pm$ 0.3 &  61.0 $\pm$ 0.8 & 59.5 $\pm$ 2.8 &  42.7 $\pm$ 1.4 & 34.3 $\pm$ 0.7 & \textbf{53.3} $\pm$ 0.4 & 98.3 $\pm$ 0.7 \\
        TCN &  35.8 $\pm$ 0.6 & 58.9 $\pm$ 0.3 &  60.2 $\pm$ 1.1 & 59.8 $\pm$ 2.8 & 41.6 $\pm$ 2.3 & 36.6 $\pm$ 0.3 & 51.8 $\pm$ 0.6 & 97.8 $\pm$ 0.9 \\
        LightGBM & 38.8 $\pm$ 0.2 & \textbf{60.4} $\pm$ 0.2 & \textbf{62.6} $\pm$ 0.1 & 57.0 $\pm$ 0.3 & 45.8 $\pm$ 2.0 & 37.1 $\pm$ 0.3 & 48.2 $\pm$ 0.4 & 99.7 $\pm$ 0.1 \\ \midrule

\multicolumn{6}{l}{\textbf{Step-wise Encoders}} \\
        \arrayrulecolor{lightgray}\midrule\arrayrulecolor{black}
        Backbone & 36.6 $\pm$ 0.5 & 59.5 $\pm$ 0.4 & 60.1 $\pm$ 0.3 & 59.3 $\pm$ 0.6 & 42.7 $\pm$ 0.3 & 31.8 $\pm$ 0.4 & 52.5 $\pm$ 0.1 & 99.1 $\pm$ 0.4 \\
        + linear embedding & 39.1 $\pm$ 0.4 & 60.5 $\pm$ 0.2 & 61.0 $\pm$ 0.8 & 58.0 $\pm$ 0.4 & 43.4 $\pm$ 1.8 & 34.5 $\pm$ 0.4 & 51.2 $\pm$ 0.8 & 97.9 $\pm$ 0.1 \\
        + MLP embedding & 38.8 $\pm$ 0.3 & 60.6 $\pm$ 0.3 & 60.7 $\pm$ 0.6 & 56.9 $\pm$ 1.1 & 41.0 $\pm$ 3.1 & 34.7 $\pm$ 0.5 & 49.6 $\pm$ 1.9 & 97.3 $\pm$ 0.3 \\
        + ResNet embedding & 37.0 $\pm$ 0.5 & 59.1 $\pm$ 0.1 & 59.2 $\pm$ 0.7 & 57.3 $\pm$ 0.7 & 43.3 $\pm$ 2.5 & 33.6 $\pm$ 0.5 & 51.5 $\pm$ 0.8 & 99.6 $\pm$ 0.5 \\
         + FTT embedding & 38.8 $\pm$ 0.6 & 59.8 $\pm$ 0.1 & 60.5 $\pm$ 0.6 &  55.7 $\pm$ 0.1 &  39.8 $\pm$ 2.6 & \textbf{38.7} $\pm$ 0.3 & 51.2 $\pm$ 0.8 & 96.9 $\pm$ 0.8 \\
         
+ FTT: type groups & \textbf{40.2} $\pm$ 0.4 & \textbf{60.3} $\pm$ 0.3 & {61.6} $\pm$ 1.0 & 54.4 $\pm$ 0.3 &  43.6 $\pm$ 0.8 & 38.0 $\pm$ 0.4 & 52.1 $\pm$ 0.1 & 97.0 $\pm$ 1.0 \\
         + FTT: organ groups  & \textbf{40.6} $\pm$ 0.4 & \textbf{60.7} $\pm$ 0.5 & \textbf{62.3} $\pm$ 1.9 & \textbf{54.0} $\pm$ 0.1 & \textbf{46.5} $\pm$ 0.6 & 37.4 $\pm$ 0.1 &  52.6 $\pm$ 0.6 & \textbf{96.4} $\pm$ 0.4 \\
        \bottomrule
        \end{tabular} }
    
\end{table*}
 \section{Experimental setup}\label{sec:exp_setup}
\setcounter{footnote}{0}
To ensure reproducibility we share our code.\footnote{\url{https://github.com/ratschlab/clinical-embeddings}}

\paragraph{Clinical prediction tasks}
We demonstrate the effectiveness of our embedding methods for electronic health records by studying their effect on prediction performance for different clinical tasks related to intensive care. Our method and related baselines are benchmarked on the online binary prediction task of (1) circulatory and (2) respiratory failure within the next 12 hours, (3) remaining length of stay and on prediction of (4) patient mortality at 24 hours after admission, as well as (5) patient phenotyping after 24 hours. Tasks (1-5), as defined in HiRID-ICU-Benchmark~\citep{yeche2021}, are based on the publicly available HiRID dataset~\citep{faltys2021hirid, hyland2020}.
We also consider the task of continuously predicting mortality \emph{within} 24 hours, throughout the patient stay -- also known as (6) {decompensation}, (7) patient mortality at 48 hours after admission and (8) remaining length of stay. We study the latter three task on the well-known MIMIC-III dataset~\citep{johnson2016}. Further details on the definition of each task can be found in benchmark papers which introduced them~\citep{harutyunyan2019multitask, yeche2021}. Further details on task definition and data pre-processing are provided in~\appendixref{appendix:tasks}.

\paragraph{Success metrics}
Our primary success metric for the usefulness of our method is \emph{performance on downstream clinical tasks}. As these often consist of significantly imbalanced classification problems \citep{yeche2021}, performance is measured through the area under the precision-recall curve (AUPRC), the area under the receiver operating characteristic curve (AUROC), and balanced accuracy (Bal. Acc.). For regression problems we report mean absolute error (MAE) in hours. This follows established practice on clinical early prediction tasks~\citep{yeche2021, harutyunyan2019multitask}. 
\paragraph{Benchmarked methods} 
We evaluate different embedding architectures including linear mapping and Feature Encoders, as referenced in ~\sectionref{sec:mod_emb}. We also compare these to deep learning models that do not use an embedding layer, where a sequential model gets the raw feature vector at each time-step. Additionally, we consider a Gradient Boosted Tree method using LightGBM~\citep{ke2017lightgbm}, based on manually-extracted features~\citep{yeche2021}. Downstream, we use deep learning backbones and optimized hyperparameters for our specific prediction tasks, as per prior research~\citep{yeche2021, harutyunyan2019multitask}. We use a Gated Recurrent Unit (GRU)~\citep{cho2014learning} network for circulatory failure prediction and a Transformer~\citep{vaswani2017attention} for all other tasks. That architectural choice for each task is based on previously published papers~\citep{yeche2022temporal, yeche2021} Further implementation details are provided in~\appendixref{appendix:implementation}.

\begin{table*}[htbp]
\floatconts
 {tab:ablation}
 {\caption{\textbf{Benchmarking analysis of embedding design choices} for circulatory failure prediction. Ablations on the default architecture: FTT \citep{gorishniy2021revisiting} with organ splitting and attention-based aggregation.}\label{tab:ablation_ftt}\vspace{-0.5cm}}
 {\subtable[Embedding architecture.][b]{\label{tab:emb_arch}
     \begin{tabular}{lc}
        \toprule
        Architecture & AUPRC \\
        \midrule
             None &  36.6 $\pm$ 0.5\\
MLP & 37.6 $\pm$ 0.8\\
             ResNet & 37.0 $\pm$ 0.5\\
             FTT  & \textbf{40.6} $\pm$ 0.4\\
         \bottomrule
    \end{tabular}
   }\qquad
   \subtable[Group aggregation.][b]{\label{tab:emb_aggr}\begin{tabular}{lc}
        \toprule
        Aggregation & AUPRC \\
        \midrule
             Concatenate & 39.4 $\pm$ 0.2 \\
             Average & 38.7 $\pm$ 0.4 \\
             Attention & \textbf{40.6} $\pm$ 0.4\\
         \bottomrule
    \end{tabular}
   }\qquad
   \subtable[Feature grouping strategies as defined in~\sectionref{sec:notation}.\vspace{4pt}][b]
   {\label{tab:group_choice}\begin{tabular}{lc}
        \toprule
        Grouping & AUPRC \\
        \midrule
        None & 38.8 $\pm$ 0.6\\
             Variable type &  39.6 $\pm$ 0.1\\
Meas. type & 39.9 $\pm$ 0.1\\
             Organ & \textbf{40.6} $\pm$ 0.4\\
         \bottomrule
    \end{tabular}
   }
 }
\end{table*} 
\section{Results}\label{sec:res}
In this section, we provide results for the proposed benchmarking study, systematically evaluating the performance of different embedding modules  for EHR modeling. We validate the following hypotheses: (1) Relying on deep learning for tabular data methods in time-step embeddings can significantly improve the performance of deep learning models for clinical time-series. (2) Specifically, via a comprehensive examination of time-step encoder components, we demonstrate that relying on the FTT approach coupled with feature grouping and appropriate aggregation tends to yield the best overall performance. (3) Attention-based embedding architectures help us gain interpretability on the feature and medical concept level into deep learning models for tabular time-series, which remains largely unexplored in the relevant literature. (4) Models based on strong clinical priors such as feature assignments to organ systems, show superior performance.

\paragraph{Advancing deep learning approaches with step-wise embedding}

First, we present the experimental results that demonstrate the performance improvement achieved through well-designed embedding methods in deep-learning models for clinical time-series predictions. Despite deep learning models\citep{cho2014learning, vaswani2017attention} often falling behind classical methods like gradient-boosted trees, as shown in~\Cref{tab:performance} and in related work~\citep{yeche2021}, we found that using tabular data deep learning techniques such as FT-Transformer~\citep{gorishniy2021revisiting} helps bridge this performance gap. Building on these insights, our proposed approach of incorporating feature grouping into the embedding process yields further significant performance gains, enabling us to overcome or match the performance of tree-based methods. We refer the interested reader to~\appendixref{appendix:extra_results} for additional results on other metrics and comparison with other methods, which further support our conclusions. Overall, our analysis establishes \textbf{a new state-of-the-art benchmark for clinical time-series tasks}, marking a substantial advancement in the field. Indeed, leveraging well-designed embedding methods and incorporating feature grouping improves performance by a similar scale to optimising the backbone architecture of sequence models in \citet{yeche2021}. 

\paragraph{Performance discrepancy between HiRID and MIMIC-III datasets.}
Incorporating feature groups within the embedding layers shows notable differences in performance gains between the HiRID and MIMIC-III datasets. This discrepancy could be attributed to two primary factors: (1) data resolution and the (2) number of features. With HiRID data resolution being twelve times greater, this leads to sequences of 2016 steps (equivalent to one week) for online tasks. The HiRID dataset processing from \citet{yeche2021} has a much greater number of features (231), compared to 18 features extracted by \citet{harutyunyan2019multitask} in MIMIC-III Benchmark. Consequently, FTT models utilizing feature grouping exhibit superior performance on the HiRID dataset, enhancing feature interaction within semantically related groups and rendering the models more resistant to noise, thereby boosting performance. Our results suggest that the use of an embedding module enables deep learning models to extracted relevant signals more effectively. On the contrary, the limited number of features available in MIMIC-III does not allow for significant performance gains with grouping, suggesting that this strategy may not be as beneficial in low-dimensional datasets.

\paragraph{Step-wise encoder with feature grouping ablation study.}
To better understand the impact of each component introduced in \Cref{sec:methods}, we investigate various design choices for step-wise embedding architectures, and analyze their impact on performance. \tableref{tab:ablation} summarizes our findings of possible concept-level architectures, feature groupings definitions, and aggregation methods. We focus on results for the circulatory failure prediction task for conciseness (referring the reader to~\appendixref{appendix:extra_results} for exhaustive results on other tasks). In \tableref{tab:emb_arch}, we find that FTT yields the largest performance gains amongst group encoder architectures. Similarly, in \tableref{tab:emb_aggr}, we find attention-based aggregation~\citep{vaswani2017attention} to consistently improve over other aggregation methods, confirming the need to capture complex concept dependencies present at a time-step level. This supports results from the tabular deep learning literature~\citep{gorishniy2021revisiting}. In addition to improving performance, we note that attention mechanisms also provide significant advantages in terms of model interpretability, as further discussed in~\sectionref{sec:interpretability}. Finally, with respect to different group definitions, we observe pre-defined grouping using domain knowledge to yield the best performance.

\begin{figure*}[htbp]
\floatconts
  {fig:attention_modality}
  {\caption{Interpretability Analysis of attention-based embeddings for respiratory failure prediction, highlighting the importance of relevant pulmonary variables -- particularly close to event occurrence.}}
  {\subfigure[Within concept embedding.]{\label{fig:attention_within}\vtop to 14em{\null\hbox{\includegraphics[width=0.315\textwidth]{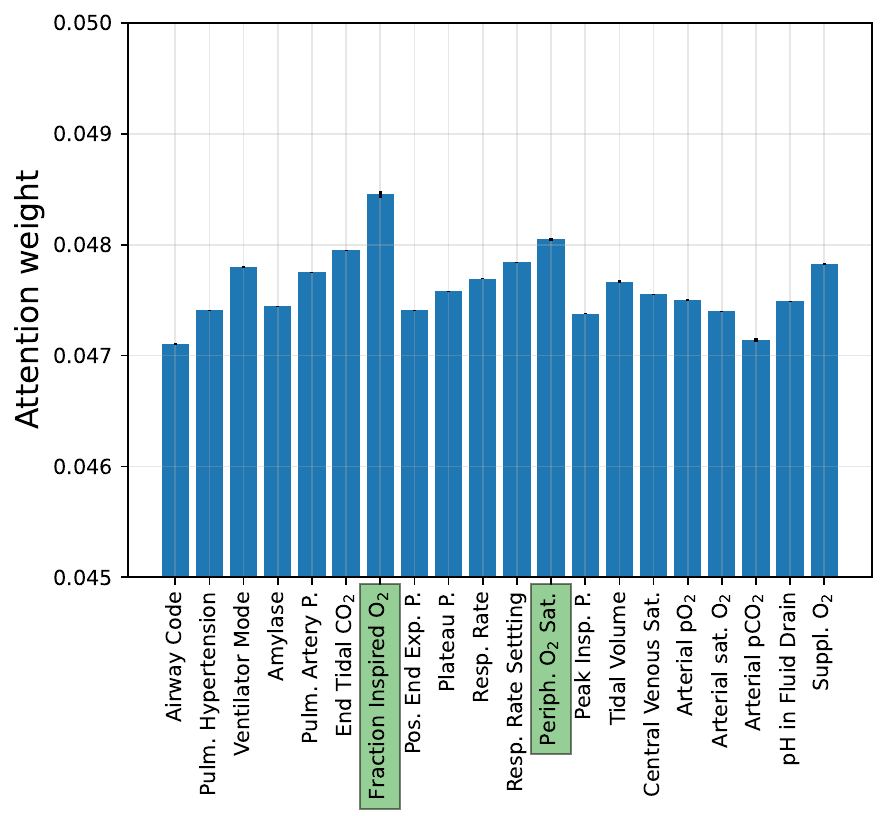}}\vfill}}\quad
    \subfigure[Between concept embeddings.]{\label{fig:attention_between}\vtop to 14em{\null\hbox{\includegraphics[width=0.315\textwidth]{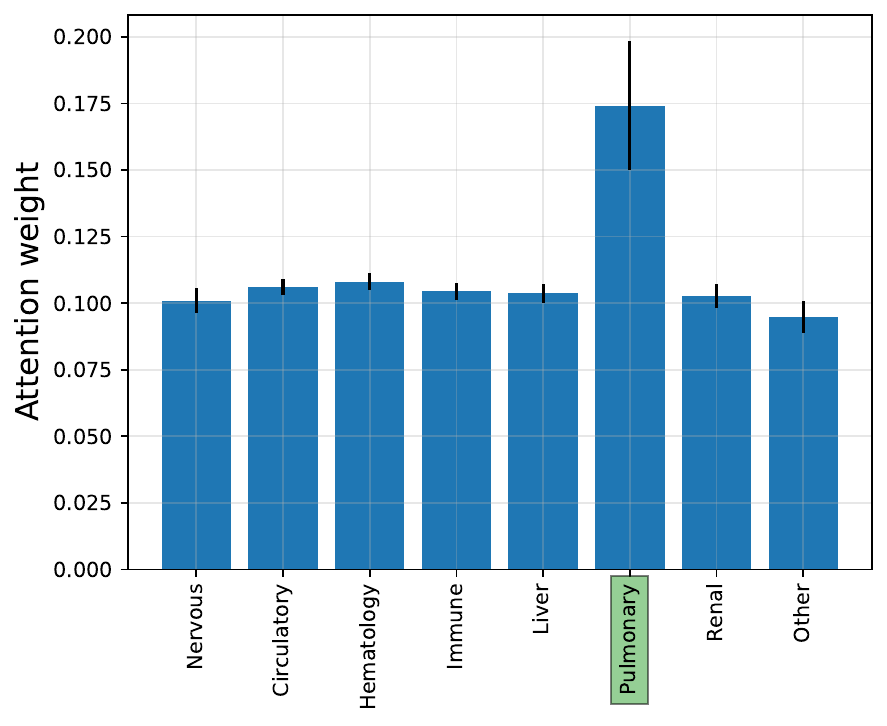}}\vfill}}
    \quad
    \subfigure[As a function of time.]{\label{fig:attention_time}\vtop to 14em{\null\hbox{\includegraphics[width=0.315\textwidth]{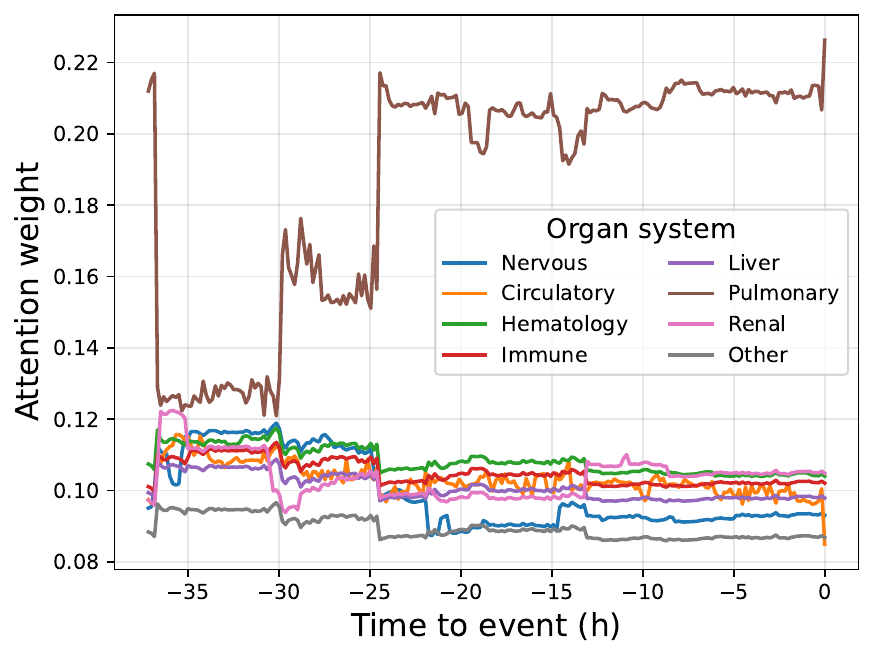}}\vfill}}
  }
\end{figure*}

\paragraph{Interpretability} \label{sec:interpretability}

As a final experiment, we explore the interpretability gained from attention-based models \citep{edward-choi-retain-neurips-2016, vig-belinkov-2019-analyzing, vig-2019-multiscale-attention} by analyzing attention at different levels of the model architecture. This provides insights into the relevance of features \textit{within} a single concept embedding as well as the differences in importance \textit{between} concept embeddings to the overall downstream prediction model. Temporal aggregations of attention scores can highlight patient trends in a given time window~\citep{cardiovascular-patient-journal-bioinformatics-gandin-2021, temporal-fusion-transformer-interpretable-att-lim-science-2021}.

In the context of respiratory failure prediction, we average attention weights over all test patient trajectories and over all timesteps. Within the group of features pertaining to the pulmonary system, we find in \figureref{fig:attention_within} that attention is on average highest on two input features that are highly predictive of this type of organ failure based on its definition~\citep{faltys2021hirid, yeche2021}: fraction of inspired oxygen (FiO$_2$) and peripheral oxygen saturation (SpO$_2$). We also find in \figureref{fig:attention_between} that the pulmonary organ system has very high importance in predicting respiratory failure, confirming that variables related to lung function, oxygen saturation and ventilation settings are correctly identified as key indicators of event imminence within the embedding model. Note that this analysis is independent of the actual label for a patient at a given time, and thus measures the average contribution of different features and groups in predicting respiratory failure. The various levels of importance scoring can be of assistance to clinicians in  different decision-making processes. For instance, on a concept level (e.g. organ systems), it can help to categorize patients in a dynamic way and make it easier to plan resources (e.g. patients with respiratory problems may require ventilators). Additionally, more detailed information on a feature level can be used to make treatment decisions.

Attention scores may not be a perfect explanation~\citep{serrano-smith-2019-attention-interpetable}, yet they can still effectively point out important signals. In a clinical decision-support context, these explanations do not need to be taken as absolute truth, but rather as a way to direct the clinician's attention to the areas that require the most care.

Another form of useful insight gained from attention-based embeddings on clinical time-series consists of patterns of attention as a function of time~\citep{temporal-fusion-transformer-interpretable-att-lim-science-2021}, as illustrated in~\figureref{fig:attention_time}. For this analysis, we plot attention weights as a function of time for individual patients within the test set. Upon entry to the intensive care unit, the attention mechanism focuses initially on the most relevant organ system,
as little patient information is available to predict imminent organ failure. As more information is acquired, attention becomes more balanced across organs, and focuses again on the pulmonary system as a respiratory failure event becomes more likely. This temporal attention pattern highlights the relevance of recent measurements and changes in variables, allowing for a deeper understanding of the predictive patterns and potential early warning signs. We refer the reader to~\appendixref{appendix:extra_results} for an exhaustive overview of this interpretability analysis, and note that this promising result could benefit from further investigations beyond the scope of the present benchmark, to correlate attention patterns with medical insights and patient evolution trends.

Overall, this study suggests that attention-based embeddings (at different levels in the architecture: features, groups, time) enhance the interpretability of deep learning models for tabular times-series, by shedding light on the most relevant features, medical concept groups and time windows for specific predictions. By understanding which variables are weighted more heavily in the model's decision-making process, clinicians and domain experts can gain trust and validate the machine learning models developed using such embedding methods \citep{ahmad2018interpretable}.

\section{Limitations \& Broader Impact}
\paragraph{Limitations} While the FT-Transformer and the use of feature groups provide 
a powerful setup for the step-wise embedding module,  it is crucial to address certain limitations associated with each.

The FT-Transformer is resource-intensive, demanding substantially more hardware and time for training compared to simpler models (MLP, ResNet, especially gradient boosting methods). Scaling it to ICU datasets like HiRID with a large number of features is challenging. Hence, the extensive use of the FT-Transformer for such datasets might increase CO2 emissions from ML pipelines. The research community has already devised a diverse range of solutions aimed at enhancing the speed, memory, and computational efficiency of Transformer-based architectures \citep{tay2022efficient}. However, when deploying actual models based on these benchmarked architectures, the performance impact of such efficiency-focused modifications remains to be explored.

On the other hand, the concept of feature groups introduces its own set of challenges. Using predefined feature groups, like organ or measurement type, may streamline the model's task, but it could limit its flexibility and requires clinical understanding for effective definition. The role of healthcare professionals is crucial for defining initial feature groups. Limitations also include the challenge of assigning each variable to a single concept, which may not fully capture the multifaceted nature of clinical data. 

\paragraph{Broader Impact}Integrating a step-wise embedding module with feature groups for ICU models, could impact both the medical and research communities. Firstly, feature grouping approach could help for \textbf{precision medicine}. By analyzing a patient's data within the context of their specific feature group, clinicians can better tailor treatment plans to address individual needs. From a machine learning viewpoint, feature splitting can amplify the \textbf{performance of predictive models}. Meaningfully grouping data permits these models to discern more complex and nuanced relationships between features, resulting in more accurate predictions. Further, the grouping of data can assist in \textbf{continuous patient monitoring}. Healthcare professionals can promptly identify any substantial changes in a patient's condition. Moreover, it can aid in assessing the risk of developing specific conditions, allowing for timely preventative measures. Finally, the \textbf{interpretability} derived from attention-based models offers enhanced \textbf{trust, validation, and transparency}. By identifying the most relevant features and feature groups and comprehending temporal dynamics through attention patterns, these models become more explainable and trustworthy. 

\section{Conclusion}

Our work benchmarks embedding architectures for deep learning as a new paradigm for clinical time-series tasks, which finally surpasses traditional tree-based methods in terms of performance. Relying on deep learning for tabular data methods, we systematically study different design choices for embedding architectures, demonstrating their essential roles in achieving state-of-the-art results. We find that for distinct groups of features predictive performance significantly improves. We also find that attention-based embeddings offer the best performance as well as greater interpretability, by identifying relevant features and feature groups -- such transparency is critical to building trust for real-world clinical applications \citep{ahmad2018interpretable}.

Overall, our study advances the field of machine learning for clinical time-series by leveraging methods and design choices from the tabular deep learning literature. We believe our findings will encourage further work in embedding design for clinical time-series, with the potential to better support clinical decision-making and improve patient outcomes.

\acks{
    This project was supported by grant \#2022-278 of the Strategic Focus Area “Personalized Health and Related Technologies (PHRT)” of the ETH Domain (Swiss Federal Institutes of Technology) and by ETH core funding (to G.R). This publication was made possible by an ETH AI Center doctoral fellowship to AP.
}

\paragraph*{Institutional Review Board (IRB)}
This research does not require IRB approval in the country in which it was performed.

\clearpage
\bibliography{kuznetsova23}

\clearpage
\appendix

\section{Clinical datasets and prediction tasks}\label{appendix:tasks}
\subsection{Task definition}
In this section, we provide more details on the definition of tasks for circulatory failure, respiratory failure and mortality from HiRID benchmark~\citep{yeche2021} and decompensation and mortality from MIMIC-III benchmark~\citep{harutyunyan2019multitask}.
 The details about the MIMIC-III and HiRID datasets, including the number of patients, endpoint definition, and statistics on annotated failure events and labels, are available in the corresponding papers that introduced these datasets: \citep{johnson2016} for MIMIC-III and \citep{faltys2021hirid} for HiRID. The respective patient splits are provided in the corresponding benchmark papers: \citep{harutyunyan2019multitask} for MIMIC-III and \citep{yeche2021} for HiRID.

HiRID benchmark tasks:
\begin{enumerate}
    \item \textbf{Circulatory failure} is a failure of the cardiovascular system, detected in practice through elevated arterial lactate ($> 2$ mmol/l) and either low mean arterial pressure ($< 65$ mmHg) or administration of a vasopressor drug. \citet{yeche2021} defines a patient to be experiencing a circulatory failure event at a given time if those conditions are met for $2/3$ of time points in a surrounding two-hour window. Binary classification, dynamic prediction throughout stay

 \item  \textbf{Respiratory failure} is defined by \citet{yeche2021} as a P/F ratio (arterial pO$_2$ over FIO$_2$) below $300$ mmHg. This definition includes mild respiratory failure. As above, \citet{yeche2021} consider a patient to be experiencing respiratory failure if $2/3$ of timepoints are positive within a surrounding 2h window. Binary classification, dynamic prediction throughout stay

 \item  \textbf{Mortality} refers to the death of the patient. The label of the time-point 24 hours after ICU admission was set to $1$ (positive) if the patient died at the end of the stay according to this field, and 0 (negative) otherwise, defining a binary classification problem to be solved once per stay. If the admission was shorter than 24 hours, no label
was assigned to the patient.

\item \textbf{Patient phenotyping} is classifying the patient after 24h regarding the  admission diagnosis, using the APACHE group II and IV labels\footnote{APACHE II and IV \citep{Zimmerman2006-of, Knaus1985-iw} are subsequent versions of the major illness severity score used in the ICU. They also introduce a patient grouping according to admission reason. We use an aggregate of these two groupings for this task (see also \citet{yeche2021})}.

\item \textbf{Remaining length of stay} is a regression task, continuous prediction of the remaining ICU stay duration.
\end{enumerate}
MIMIC-III benchmark tasks:
\begin{enumerate}
    \item \textbf{Decompensation} refers to the death of a patient in the next $24$h. The event labels are directly extracted from the MIMIC-III \citep{johnson2016} metadata about the time of death of a patient. 

\item \textbf{Mortality} refers to the death of a patient after $48$ hours of observed ICU data. The event labels are directly extracted from the MIMIC-III \citep{johnson2016} metadata.

\item \textbf{Length of stay} is a prediction of the remaining time the patient will stay in the ICU.
\end{enumerate}

MIMIC-III license is \href{https://physionet.org/content/mimiciii/view-license/1.4/}{PhysioNet Credentialed Health Data License 1.5.0}; HiRID~--- \href{https://physionet.org/content/hirid/view-license/1.1.1/}{PhysioNet Contributor Review Health Data License 1.5.0}.

\subsection{Pre-processing}

We describe the pre-processing steps we applied to both datasets, HiRID and MIMIC-III.

\paragraph{Imputation.}
Diverse imputation methods exist for ICU time series. For simplicity, we follow the approach of original benchmarks \citep{harutyunyan2019multitask,yeche2021} by using forward imputation when a previous measure existed. The remaining missing values are zero-imputed after scaling, corresponding to a mean imputation. 

\paragraph{Scaling.} Whereas prior work explored clipping the data to remove potential outliers \citep{tomavsev2019clinically}, we do not adopt this approach as we found it to reduce performance on early prediction tasks. A possible explanation is that, due to the rareness of events, clipping extreme quantiles may remove parts of the signal rather than noise. Instead, we simply standard-scale data based on the training sets statistics. 

\section{Implementation details} \label{appendix:implementation}

\begin{table}[tbh!]
\caption{\textbf{Hyperparameter search range} for {mortality, MIMIC-III} with Transformer\citep{vaswani2017attention} backbone. In \textbf{bold} are parameters selected by random search.}
\begin{tabular}{lc}
\toprule
Hyperparameter & Values\\
\midrule
\midrule
Learning Rate & (1e-5, 3e-5, \textbf{1e-4}, 3e-4) \\
\midrule
Drop-out & (0.0, 0.1, 0.2, {0.3}, \textbf{0.4}) \\
\midrule
Attention Drop-out &   (0.0, {0.1}, 0.2, \textbf{0.3}, 0.4) \\
\midrule
Depth &   (\textbf{1}, {2}, 3) \\
\midrule
Heads &  (\textbf{1}, \textbf{2}, 4) \\
\midrule
Hidden Dimension &  (16, 32, \textbf{64}) \\
\midrule
L1 Regularization &  (\textbf{1e-3}, 1e-2, {1e-1}, 1, 10)\\
\bottomrule
\end{tabular}
\label{tab:hp-search-ihm}
\end{table}

\begin{table*}[tbh!]

    \centering
\begin{tabular}{l|c|c|c}

\toprule
Task & Depth &  Colsample\_bytree \tablefootnote{Subsample ratio of columns when constructing each tree.} & Subsample \tablefootnote{Subsample ratio of the training instance} \\
\midrule
\midrule
Mortality & (3, \textbf{4}, 5, 6, \textbf{7})  & (0.33, \textbf{0.66}, 1.00) & (0.33, 0.66, \textbf{1.00})\\
Decompensation & (3, 4, 5, \textbf{6}, 7) & (\textbf{0.33}, 0.66, 1.00)& (0.33, 0.66, \textbf{1.00}) \\
\bottomrule
\end{tabular}
    \caption{Hyperparameter search range for LGBM. In \textbf{bold} are the parameters we selected using random search.}
    \label{tab:hp-search-gb}
\end{table*}

\subsection{Modality splitting} \label{appendix:splittings}
Organ splitting is detailed in \tableref{tab:organ_splitting_h} and \tableref{tab:organ_splitting_m}. Splitting by variable type is provided in \tableref{tab:type_splitting_h} and \tableref{tab:type_splitting_m}. Both are obtained from metadata in HiIRID and MIMIC-III datasets, which specify which organ or value type each variable belongs to. Measurement splitting is determined by whether the variable is numerical or categorical, and this can be found in the related dataset descriptions~\citep{faltys2021hirid, johnson2016}.

\subsection{Training Setup}
\label{appendix: setup}

\paragraph{Training details.} For all models, we set the batch size according to the available hardware capacity. We use \texttt{Nvidia RTX2080 Ti} GPUs with 11GB of GPU memory. Depending on the model size, dataset and task, we use between 1 to 8 GPUs in a distributed data-parallel mode. We early stopped each model training according to their validation loss when no improvement was made after 10 epochs. 

\paragraph{Libraries.} A full list of libraries and the version we used is provided in the \texttt{environment.yml} file. The main libraries on which we build our experiments are the following: pytorch 1.11.0 \citep{NEURIPS2019_9015}, scikit-learn 0.24.1\citep{scikit-learn}, ignite 0.4.4, CUDA 10.2.89\citep{cuda}, cudNN 7.6.5\citep{chetlur2014cudnn}, gin-config 0.5.0 \citep{gin}.

\paragraph{Infrastructure.}
We follow all guidelines provided by \texttt{pytorch} documentation to ensure the reproducibility of our results. However, reproducibility across devices is not ensured. Thus we provide here the characteristics of our infrastructure. We trained all models on a 1 to 8 \texttt{Nvidia RTX2080 Ti} with a \texttt{Xeon E5-2630v4} CPU. 
Training took between 3 and 10 hours for a single run.

\paragraph{Architecture choices for the sequential backbone model.} We used the same architecture and hyperparameters reported giving the best performance on circulatory failure, respiratory failure and decompensation in \citet{yeche2022temporal}. For all other tasks from HiRID benchmark, we used the same architecture and hyperparameters reported in \citet{yeche2021}. For mortality, MIMIC-III benchmark we carried out our own random search on validation AUPRC performance. The exact parameters for this task are reported in ~\tableref{tab:hp-search-ihm}. 

\paragraph{Gradient Boosting} 
We used the same architecture and hyperparameters reported giving the best performance on HiRID benchmark tasks in \citet{yeche2021}. For mortality and decompensation, MIMIC-III benchmark we carried out our own random search on validation AUPRC performance. The range of hyperparameters considered for the gradient boosting method, LightGBM framework\footnote{\url{https://lightgbm.readthedocs.io/en/latest/}} can be found in \tableref{tab:hp-search-gb}.

\subsection{Embedding architectures}

We follow MLP, ResNet and FT-Transformer implementation, described in \citet{gorishniy2021revisiting}. Architecture and hyperparameters investigated for each task are given in \tableref{tab:random_search_results_resnet_mlp} for MLP and ResNet architectures and in \tableref{tab:random_search_results_ftt} for FT-Transformer, along with the setting giving optimal validation performance in each case.

\begin{table*}[tbh!]
    \centering
    \caption{\textbf{Embedding architecture and hyperparameter values} studied for each clinical prediction task for MLP and ResNet architectures. Best values, obtained by random search over the proposed grid, are highlighted in bold.}
    \label{tab:random_search_results_resnet_mlp}
    \resizebox{\textwidth}{!}{\begin{tabular}{lccccc}
    \toprule
    Dataset & \multicolumn{3}{c} {HiRID} & \multicolumn{2}{c}{MIMIC-III} \\
    \cmidrule(lr){2-4} \cmidrule(lr){5-6}
    Clinical prediction task & Circulatory Failure & Respiratory Failure & Mortality & Decompensation & Mortality \\ \midrule
Embedding architecture & (\textbf{MLP}, ResNet) & (\textbf{MLP}, ResNet) & (\textbf{MLP}, {ResNet}) & (\textbf{MLP}, ResNet) & (\textbf{MLP}, ResNet) \\
        Modality split & (none, organ, categorical, \textbf{type}) & (none, \textbf{organ}, categorical, type) & (none, \textbf{organ}, \textbf{categorical}, type) & (none,  \textbf{organ}, categorical, type) & ( \textbf{none}, organ, categorical, type) \\
         Aggregation & (avg., \textbf{concat}., attention) & (avg., \textbf{concat.}, attention) & (\textbf{avg.}, {concat.}, attention) & (avg., concat., \textbf{attention}) & ( \textbf{avg.}, concat., attention) \\
         Embedding depth & (1 2 \textbf{3} 4) & (1 \textbf{2} 3 4) & (\textbf{1} {2} 3 4) & (1 2 \textbf{3} 4) & ( \textbf{1} 2 3 4)\\
         Embedding latent dim. & (8 16 \textbf{32} 64) & (8 16 \textbf{32} 64) & (8 {16} 32 \textbf{64}) & (8 16 32 \textbf{64}) & (8 16  \textbf{32} 64)\\
         L1 regularization weight & (0 0.1 1 \textbf{10}) & (0 0.1 1 \textbf{10}) & (0 \textbf{0.1} 1 10) & (0 \textbf{0.1} 1 10) & ( \textbf{0} 0.1 1 10)\\
    \bottomrule
    \end{tabular}}
\end{table*}

\begin{table*}[tbh!]
    \centering
    \caption{\textbf{Embedding architecture and hyperparameter values} studied for each clinical prediction task for FTT architecture. Best values, obtained by random search over the proposed grid, are highlighted in bold.}
    \label{tab:random_search_results_ftt}
    \resizebox{\textwidth}{!}{\begin{tabular}{lccccc}
    \toprule
    Dataset & \multicolumn{3}{c} {HiRID} & \multicolumn{2}{c}{MIMIC-III} \\
    \cmidrule(lr){2-4} \cmidrule(lr){5-6}
    Clinical prediction task & Circulatory Failure & Respiratory Failure & Mortality & Decompensation & Mortality \\ \midrule
Modality split & (none, \textbf{organ}, categorical, type) & (none, \textbf{organ}, categorical, type) & (none, \textbf{organ}, categorical, type) & (\textbf{none}, organ, categorical, type) & (none, \textbf{organ}, categorical, type)\\
         Aggregation & (avg., concat., \textbf{attention}) & (avg. concat., \textbf{attention}) & (avg., \textbf{concat.}, attention) & (avg., concat., \textbf{attention}) & (\textbf{avg.}, concat., attention) \\
         FTT token dim & (32 \textbf{64}) & (32 \textbf{64}) & (16 32 \textbf{64} {128}) & (16 32 \textbf{64} 128) &(16 32 \textbf{64} 128)\\
         FTT depth & (\textbf{1} 2) & (\textbf{1} 2) & (\textbf{1} {2} 3) & (1 \textbf{2} 3) & (\textbf{1} 2 3)\\
FTT heads & (1 \textbf{2} 3) & (1 {2} \textbf{3}) & (1 \textbf{2} 3 ) & (1 2 \textbf{3} ) & (1 2 \textbf{3})\\
    \bottomrule
    \end{tabular}}
\end{table*}

\subsection{Concept aggregation}

Embeddings from each concept are aggregated by taking the average of the multiple embeddings, concatenating them, of computed an attention-based aggregation. Hyperparamters investigated for each task for attention-based aggregation are given in \tableref{tab:random_search_results_agg}.

\begin{table*}[tbh!]
    \centering
    \caption{\textbf{Hyperparameter values} studied for each clinical prediction task for attention-based aggregation. Best values, obtained by random search over the proposed grid, are highlighted in bold.}
    \label{tab:random_search_results_agg}
    \resizebox{\textwidth}{!}{\begin{tabular}{lccccc}
    \toprule
    Dataset & \multicolumn{3}{c} {HiRID} & \multicolumn{2}{c}{MIMIC-III} \\
    \cmidrule(lr){2-4} \cmidrule(lr){5-6}
    Clinical prediction task & Circulatory Failure & Respiratory Failure & Mortality & Decompensation & Mortality \\ \midrule
Agg. depth & (1 \textbf{2} 3) & (1 \textbf{2} 3) & (1 \textbf{2} 3 ) & (1 \textbf{2} 3 ) & (1 \textbf{2} 3)\\
Agg. heads & (1 2 \textbf{3}) & (1 2 \textbf{3} ) & (1 2 \textbf{3} ) & (1 \textbf{2} 3 ) & (\textbf{1} 2 3) \\
    \bottomrule
    \end{tabular}}
\end{table*}

\section{Additional results and ablations} \label{appendix:extra_results}

In this Section, we provide the additional results on other metrics, which support our conclusions from the \sectionref{sec:res}.

\subsection{Comparison with unsupervised pretraining techniques}
In \tableref{tab:unsupervised-pretrain-comparison} we provide a comparison with pretraining techniques followed by training MLPs on top of the pretrained representations to perform the downstream prediction tasks. For fair comparison with \citet{yeche2021neighborhood} we used a temporal convolutional network (TCN) as the backbone sequence architecture.

\begin{table*}[tb]
\floatconts
  {tab:unsupervised-pretrain-comparison}{
      \caption{Comparison with a set of unsupervised pretraining techniques on two MIMIC-III benchmark tasks: Decompensation and Length-of-stay. Semi-supervised approaches inlcude some labels to pretrain patient representations.}
    }{
        \footnotesize
\begin{tabular}{lccc}
        \toprule
        Task & \multicolumn{2}{c}{Decompensation} & Length-of-stay \\
        \cmidrule(lr){2-3} \cmidrule(lr){4-4}
        Metric & AUPRC & AUROC & Kappa \\

        \midrule
        \multicolumn{4}{l}{\textbf{Self-Supervised Pretraining}} \\
        \arrayrulecolor{lightgray}\midrule\arrayrulecolor{black}

        SACL \citep{cheng2020subject} &  29.3 $\pm$ 0.9 & 87.5 $\pm$ 0.4 & 40.1 $\pm$ 0.5 \\
        CLOCS \citep{kiyasseh2021clocs}   & 32.2 $\pm$ 0.8 & 90.5 $\pm$ 0.2 & 43.0 $\pm$ 0.2 \\
        NCL~\citep{yeche2021neighborhood}          & 35.1 $\pm$ 0.4 & 90.8 $\pm$ 0.2 & 43.2 $\pm$ 0.2 \\

        \midrule
        \multicolumn{4}{l}{\textbf{Semi-Supervised Pretraining}} \\
        \arrayrulecolor{lightgray}\midrule\arrayrulecolor{black}
        SCL (D) \citep{khosla2020supervised}      &  32.1 $\pm$ 0.9 & 89.5 $\pm$ 0.3 & 41.8 $\pm$ 0.4 \\
NCL~\citep{yeche2021neighborhood}  & 37.1 $\pm$ 0.7 & 90.9 $\pm$ 0.1 & 43.8 $\pm$ 0.3 \\

        \midrule
        \multicolumn{4}{l}{\textbf{Our Supervised Step-Wise Embedding Approach}} \\
        \arrayrulecolor{lightgray}\midrule\arrayrulecolor{black}
        FTT embedding~\citep{gorishniy2021revisiting} & \textbf{38.2} $\pm$ 0.4 & 90.9 $\pm$ 0.3 & 42.9 $\pm$ 0.6 \\
        FTT with organ grouping & \textbf{38.2} $\pm$ 0.5 & \textbf{91.1} $\pm$ 0.3 & \textbf{44.0} $\pm$ 0.3 \\
        
        \bottomrule
        \end{tabular}
    }
\end{table*}

\subsection{Additional performance benchmark results for embedding architectures}
In this section, we additionally compare the previously described methods on all tasks. First, we report AUROC metric for the results, given in the \sectionref{sec:res},~\tableref{tab:performance}, see \tableref{tab:performance_auroc}.

\begin{table*}[tb]
\floatconts
  {tab:performance_auroc}{
      \caption{\textbf{Performance benchmark for different embedding architectures,} measured through the receiver operating characteristic curve (AUROC). Mean and standard deviation are reported over five training runs.}
    }{
        \footnotesize
        \setlength{\tabcolsep}{3pt}
        \begin{tabular}{lccccc}
        \toprule
            Dataset & \multicolumn{3}{c} {HiRID} & \multicolumn{2}{c}{MIMIC-III} \\
            \cmidrule(lr){2-4} \cmidrule(lr){5-6}
            Clinical prediction task & Circulatory Failure & Respiratory Failure & Mortality & Decompensation & Mortality \\ \midrule
            LightGBM \citep{yeche2021} & \textbf{91.2} $\pm$ 0.1 & \textbf{70.8} $\pm$ 0.1 & 90.5 $\pm$ 0.0 & 90.3 $\pm$ 0.1 & 84.2 $\pm$ 0.1\\ \midrule
            Deep learning backbone (DL) & 90.5 $\pm$ 0.2 & 69.9 $\pm$ 0.4 & 90.7 $\pm$ 0.2 & 90.5 $\pm$ 0.1 & 86.1 $\pm$ 0.1 \\
            + linear embedding \citep{yeche2021} &90.9 $\pm$ 0.1 &\textbf{71.0} $\pm$ 0.2 & 90.8 $\pm$ 0.2 & 91.1 $\pm$ 0.1 &85.8 $\pm$ 0.2\\
            + MLP embedding \citep{tomavsev2021use} &91.0 $\pm$ 0.1 &70.7 $\pm$ 0.3 &90.5 $\pm$ 0.1 & 91.1 $\pm$ 0.5 & 85.6 $\pm$ 0.1\\
            + ResNet embedding \citep{tomavsev2019clinically} &90.1 $\pm$ 0.3 &69.5 $\pm$ 0.1 & 89.9 $\pm$ 0.2 & 90.7 $\pm$ 0.2 & 85.9 $\pm$ 0.2\\
             + FTT embedding \citep{gorishniy2021revisiting} & 91.1 $\pm$ 0.1& 70.0 $\pm$ 0.1 & 90.5 $\pm$ 0.2& \textbf{91.6} $\pm$ 0.1 & 85.8 $\pm$ 0.2	\\
+ FTT with type grouping & {91.0} $\pm$ 0.2 & \textbf{70.6} $\pm$ 0.2 & 90.1 $\pm$ 0.3& \textbf{91.4} $\pm$	0.1 & \textbf{86.0} $\pm$ 0.2\\
             + FTT with organ grouping  & \textbf{91.6} $\pm$ 0.03& \textbf{70.6} $\pm$ 0.4 & \textbf{91.0} $\pm$ 0.3 & \textbf{91.4} $\pm$ 0.1 & \textbf{86.1} $\pm$ 0.2 \\
            \bottomrule
        \end{tabular}
    }
\end{table*}

In addition to \tableref{tab:ablation} we summarize our findings of possible concept-level architectures, feature groupings definitions, and aggregation methods on all other tasks in \tableref{tab:ablation_resp} - \tableref{tab:ablation_ihm}.

\subsection{Interpretability}

\paragraph{Additional tasks.}

\begin{figure*}[tbh!]
    \centering
\includegraphics[width=\textwidth]{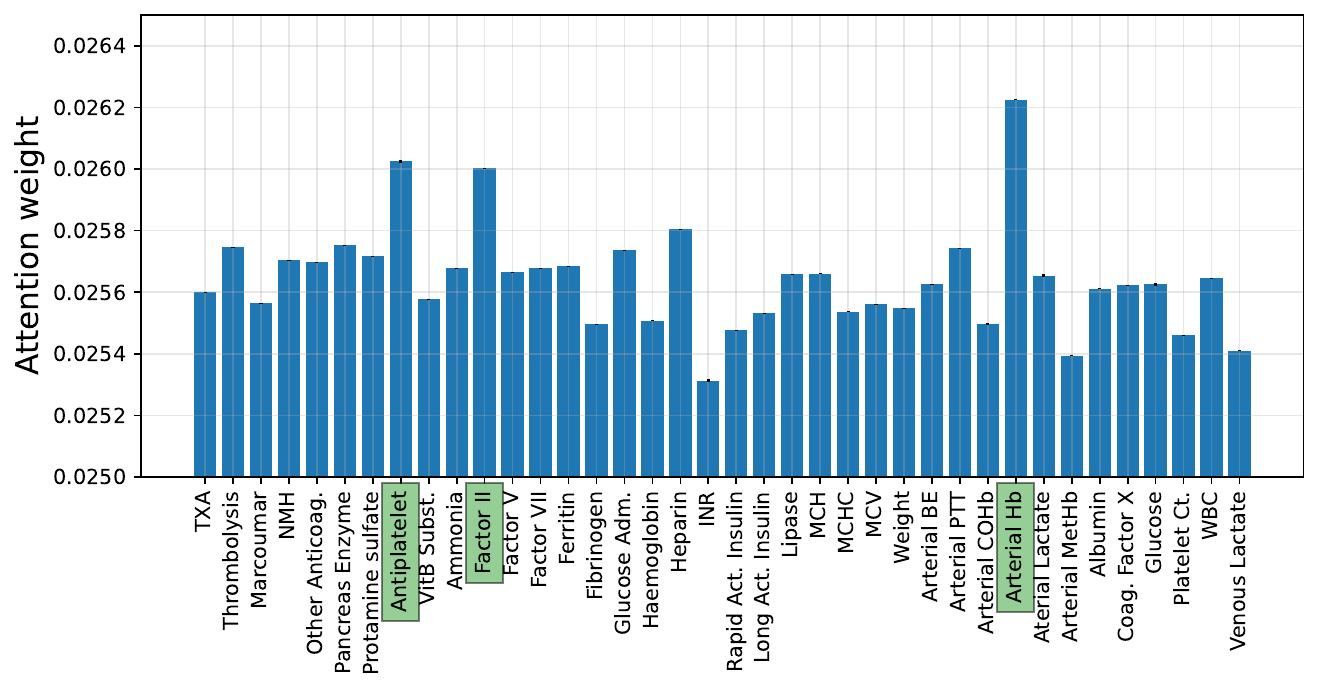}
        \caption{Within concept embedding (hematology system).}
        \label{fig:attention_between_circ}
\end{figure*}

\begin{figure*}[tbh!]
    \centering
\includegraphics[width=.7\textwidth]{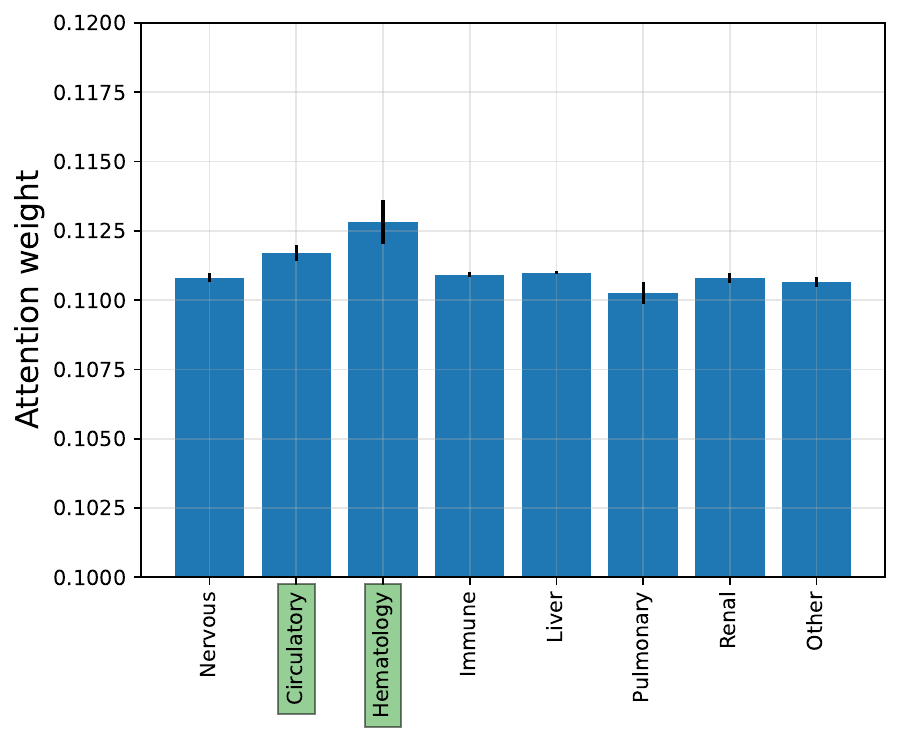}
        \caption{Between concept embeddings (organ systems).}
        \label{fig:attention_between_circ_2}
\end{figure*}

Additional results on circulatory failure prediction are shown in \figureref{fig:attention_between_circ} and in \figureref{fig:attention_between_circ_2}. Average attention weights between different organ systems, highlight the importance of relevant groups of features in predicting the corresponding organ failure. We find that the cardiovascular and hematology organ systems show the highest relevance to predicting circulatory failure, confirming that variables related to heart function, blood pressure, and vascular dynamics may play a critical role. Overall, features and organ groups with highest attention weights correspond to important predictive variables from a clinical perspective as shown by \citet{hyland2020}.

\paragraph{Attention over time.} We provide additional examples of attention pattern over time in \figureref{fig:attention_time_extra_resp} and \figureref{fig:attention_time_extra_circ}, showing the insights gained from attention-based embedding methods in interpreting model behaviour.

\section{Prior work: deep learning backbones for ICU data}\label{appendix: prior_work}

As was mentioned, the performance gap between proposed deep learning methods and tree-based approaches remains significant~\citep{yeche2021, hyland2020}. Some approaches have considered the use of additional data sources via fusion models~\citep{husmann2022importance, khadanga2019using} to achieve comparable performance. In this section, we also aim to address several prior papers that have contributed to the development of backbone model architectures for supervised clinical time-series tasks. The SeFT model~\citep{DBLP:conf/icml/HornMBRB20} treats observations as tuples of time value, observed variable value, and modality indicator. These tuples are concatenated and each individually passed through a linear layer to generate embeddings, which then are aggregated across the \emph{entire time-series}. The RAINDROP model~\citep{zhang2021graph} maps each observed variable to a high-dimensional space with an MLP and uses Graph Neural Networks to learn relevant relationships. Separate line of research explores self-supervised pre-training methodologies for clinical time-series representation learning \citep{tipirneni2022self,labach2023effective}. StraTS~\citep{tipirneni2022self} represents the data in the same way as SeFT. The TESS model~\citep{labach2023effective} considers time bins which are passed through an MLP.  
To summarize, SeFT and StraTS employ the same architecture, where the features interact within the whole time-series, and necessitates a specific data representation. Time-step level and group level interactions are not in the scope of these studies. Similar to SeFT, RAINDROP aggregates information across the entire time-series for the feature embeddings and employs an architecture not suited for online prediction tasks. TESS and StraTS focus on exploring the effects of self-supervised pre-training, which is distinct from the focus of our research.

\begin{figure*}[htbp]
\floatconts
  {fig:attention_time_extra_resp}
  {\caption{\textbf{Attention patterns over time} in embeddings for clinical time-series for Respiratory failure prediction task. Example attention weights between different organ systems.}}
  {\subfigure[]{\vtop to 14em{\null\hbox{\includegraphics[width=0.315\textwidth]{figures/attention_vs_time.pdf}}\vfill}}\quad
    \subfigure[]{\vtop to 14em{\null\hbox{\includegraphics[width=0.315\textwidth]{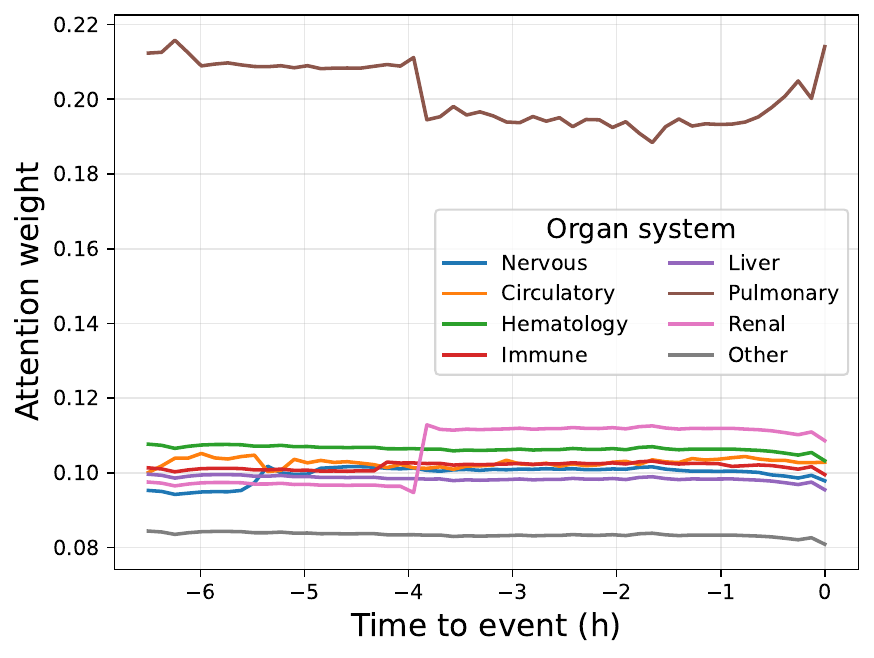}}\vfill}}
    \quad
    \subfigure[]{\vtop to 14em{\null\hbox{\includegraphics[width=0.315\textwidth]{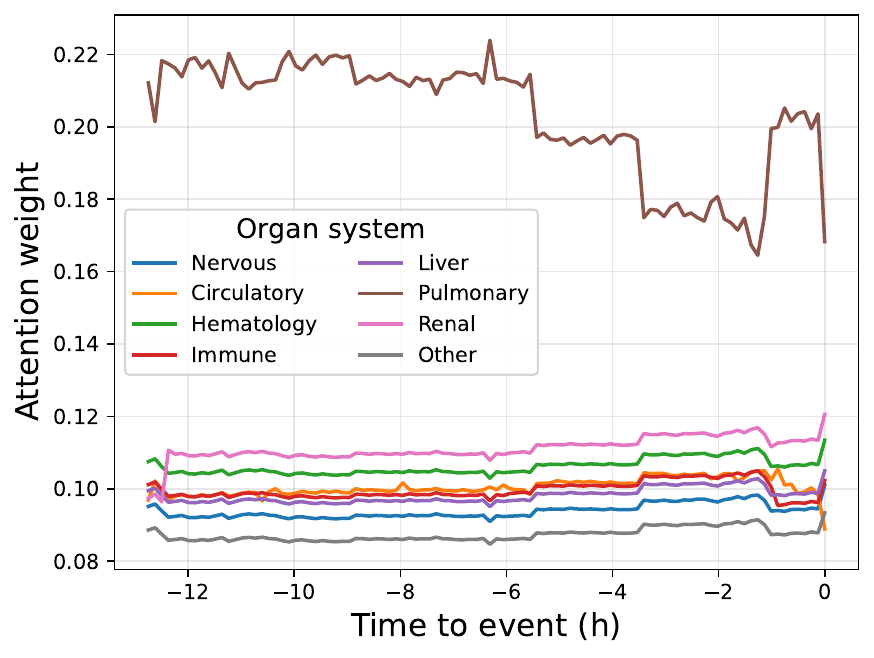}}\vfill}}
  }
\end{figure*}

\begin{figure*}[htbp]
\floatconts
  {fig:attention_time_extra_circ}
  {\caption{\textbf{Attention patterns over time} in embeddings for clinical time-series for Circulatory failure prediction task. Example attention weights between different organ systems.}}
  {\subfigure[]{\vtop to 14em{\null\hbox{\includegraphics[width=0.315\textwidth]{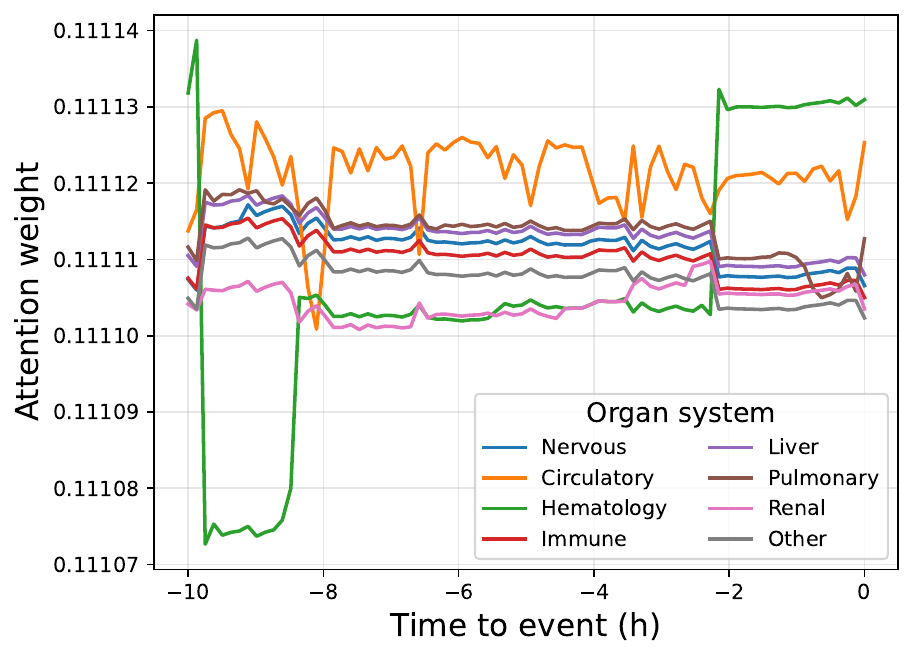}}\vfill}}\quad
    \subfigure[]{\vtop to 14em{\null\hbox{\includegraphics[width=0.315\textwidth]{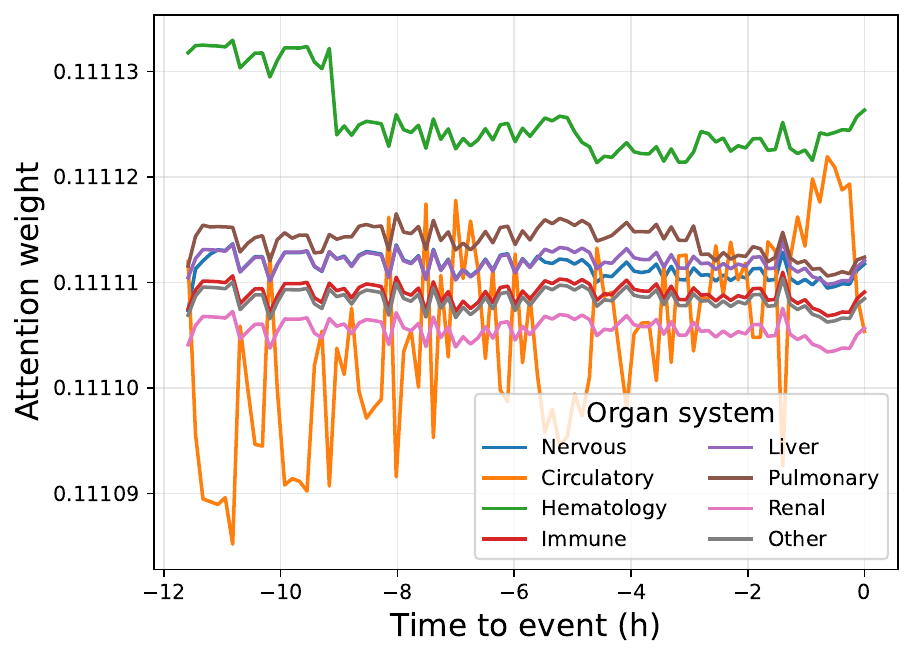}}\vfill}}
    \quad
    \subfigure[]{\vtop to 14em{\null\hbox{\includegraphics[width=0.315\textwidth]{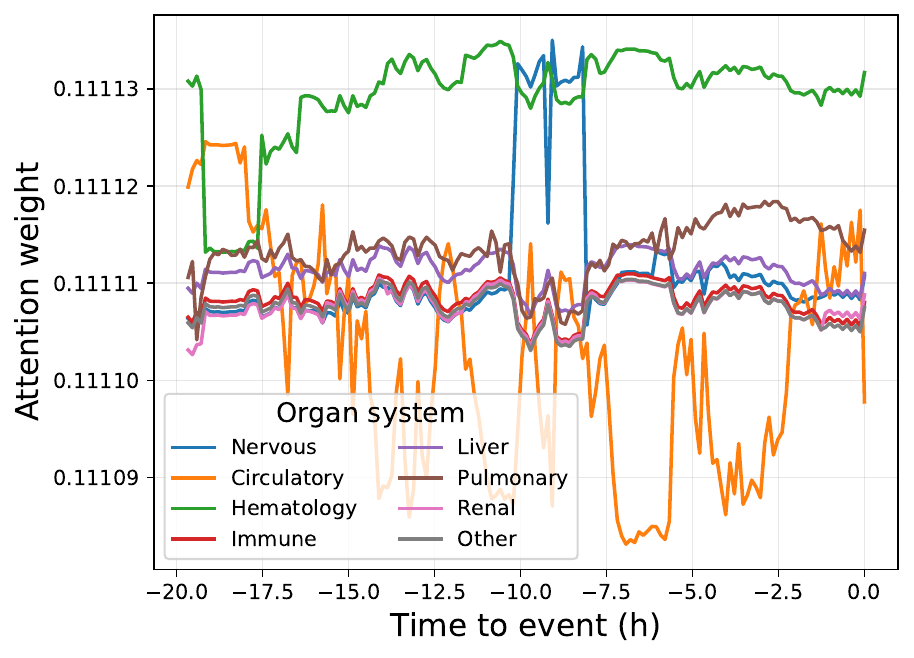}}\vfill}}
  }
\end{figure*}
\begin{table*}[tbh!]
\small 
    \centering
    \caption{\textbf{Variable splitting by organ type,} obtained based on public metadata in the HiRID dataset. An intensive care physician was consulted to confirm the validity of these splits. Details on variable name and acronyms can be obtained from the respective datasets \citep{hyland2020}.}
    \label{tab:organ_splitting_h}
    
\resizebox{\textwidth}{!}{\begin{tabular}{l p{12cm}}
    \toprule
        Organ & Variable name \\ \midrule
        \textbf{Central nervous system} (Brain) &  GCS Antwort, GCS Motorik, GCS Augenöffnen, RASS, ICP, TOF, Benzodiacepine, Alpha 2 Agonisten, Barbiturate, Propofol, Liquor/h, Nimodipin, Opiate, Non-opioide, NSAR, Ketalar, Peripherial Anesthesia, Antiepileptica, Anti delirant medi, Psychopharma, Muskelrelaxans, Anexate, Naloxon, Parkinson Medikaiton, pH Liquor, Laktat Liquor, Glucose Liquor \\
        \textbf{Circulatory system} (Heart)  &  HR, T Central, ABPs, ABPd, ABPm, NIBPs, NIBPd, NIBPm, PAPm, PAPs, PAPd, PCWP, CO, SvO2(m), ZVD, ST1, ST2, ST3, Rhythmus, IN, OUT, Incrys, Incolloid, packed red blood cells, FFP, platelets, coagulation factors, norepinephrine, epinephrine, dobutamine, milrinone, levosimendan, theophyllin, vasopressin, desmopressin, vasodilatators, ACE Inhibitors, Sartane, Ca Antagonists, B-Blocker, Andere, Adenosin, Digoxin, Amiodaron, Atropin, Thyrotardin, Thyroxin, Thyreostatikum, Mineralokortikoid, Antihistaminka, Terlipressin, Troponin-T, creatine kinase, creatine kinase-MB, BNP, TSH, AMYL-S \\
        \textbf{Hematology} (Blood)  &  Glucose Administration, Insuling Langwirksam, Insulin Kurzwirksam, Thrombozytenhemmer, Heparin, NMH, Others in Case of HIT, Marcoumar, Protamin, Anti Fibrinolyticum, Lysetherapie, Pankreas Enzyme, VitB Substitution, Weight, a-BE, a\_COHb, a\_Hb, a\_Lac, a\_MetHb, v-Lac, aPTT, Fibrinogen, FII, Factor V, Factor VII, factor X, INR, albumin, glucose, Ammoniak, Hb, total white blood cell count, platelet count, MCH, MCHC, MCV, Ferritin, Lipase \\
        \textbf{Immune system}  &  Administriation of antibiotics, Administation of antimycotic, administration of antiviral, Antihelmenticum, Steroids, Enteral Feeding, steroids, non-steroids, Chemotherapie, Immunoglobulin, Immunsuppression, GCSF, C-reactive protein, procalcitonin, lymphocyte, Neutr, Segm. Neut., Stabk. Neut., BSR, Cortisol \\
        \textbf{Hepatic system} (Liver)  &  ASAT, ALAT, bilirubine, total, Bilirubin, direct, alkaline phosphatase, gamma-GT \\
        \textbf{Pulmonary system} (Lung)  &  SpO2, ETCO2, RR, supplemental oxygen, FIO2, Peep, Ventilator mode, TV, Spitzendruck, Plateaudruck, AWPmean, RR set, AiwayCode, Beh. Pulm. Hypertonie, a\_pCO2, a\_PO2, a\_SO2, Zentral venöse sättigung, pH Drain, AMYL-Drainag \\
        \textbf{Renal system} (Kidneys)  &  OUTurine/h, K-sparend, Aldosteron Antagonist, Loop diuretics, Thiazide, Acetazolamide, Haemofiltration, Parenteral Feeding, Kalium, Phosphat, Na, Mg, Ca, Trace elements, Bicarbonate, a\_HCO3-, a\_pH, K+, Na+, Cl-, Ca2+ ionizied, Ca2+ total, phosphate, Mg\_lab, Urea, creatinine, urinary creatinin, urinary Na+, urinary urea \\
    \bottomrule
    \end{tabular}}
    \end{table*}
     \begin{table*}[tbh!]\caption{\textbf{Variable splitting by organ type,} obtained based on public metadata in MIMIC-III dataset. An intensive care physician was consulted to confirm the validity of these splits. Details on variable name and acronyms can be obtained from the respective datasets \citep{hyland2020}.}
    \label{tab:organ_splitting_m}
    \resizebox{\textwidth}{!}{\begin{tabular}{l p{12cm}}
    \toprule
        Organ & Variable name \\ \midrule
        \textbf{Central nervous system} (Brain) & Glascow coma scale eye opening, Glascow coma scale motor response, Glascow coma scale total, Glascow coma scale verbal response \\
        \textbf{Circulatory system} (Heart)  &  Diastolic blood pressure, Heart Rate, Mean blood pressure, Systolic blood pressure, Temperature, Capillary refill rate \\
        \textbf{Hematology} (Blood)  & Glucose \\
        \textbf{Pulmonary system} (Lung)  &  Fraction inspired oxygen, Oxygen saturation, Respiratory rate \\
        \textbf{Renal system} (Kidneys)  & pH \\
    \bottomrule
    \end{tabular}}
    \end{table*}
 \begin{table*}[h]
    \centering
    \caption{\textbf{Variable splitting by data acquisition type,} obtained based on public metadata in the HiRID dataset.}
    \label{tab:type_splitting_h}
    
\resizebox{\textwidth}{!}{\begin{tabular}{l p{12cm}}
    \toprule
        Variable type & Variable name \\ \midrule
\textbf{Derived from raw data}  &  ETCO2, OUTurine/h, IN, OUT, Incrys, Incolloid \\
\textbf{Laboratory values}  &  a-BE, a\_COHb, a\_Hb, a\_HCO3-, a\_Lac, a\_MetHb, a\_pH, a\_pCO2, a\_PO2, a\_SO2, Zentral venöse sättigung, Troponin-T, creatine kinase, creatine kinase-MB, v-Lac, BNP, K+, Na+, Cl-, Ca2+ ionizied, Ca2+ total, phosphate, Mg\_lab, Urea, creatinine, urinary creatinin, urinary Na+, urinary urea, ASAT, ALAT, bilirubine, total, Bilirubin, direct, alkaline phosphatase, gamma-GT, aPTT, Fibrinogen, FII, Factor V, Factor VII, factor X, INR, albumin, glucose, Ammoniak, C-reactive protein, procalcitonin, lymphocyte, Neutr, Segm. Neut., Stabk. Neut., BSR, Hb, total white blood cell count, platelet count, MCH, MCHC, MCV, Ferritin, TSH, AMYL-S, Lipase, Cortisol, pH Liquor, Laktat Liquor, Glucose Liquor, pH Drain, AMYL-Drainag \\
\textbf{Monitored variables}  &  HR, T Central, ABPs, ABPd, ABPm, NIBPs, NIBPd, NIBPm, PAPm, PAPs, PAPd, PCWP, CO, SvO2(m), ZVD, ST1, ST2, ST3, SpO2, ETCO2, RR, ICP, TOF, FIO2, Peep, Ventilator mode, TV, Spitzendruck, Plateaudruck, AWPmean, RR set \\
\textbf{Observed variables}  &  ZVD, Rhythmus, supplemental oxygen, GCS Antwort, GCS Motorik, GCS Augenöffnen, RASS, ICP, AiwayCode, Haemofiltration, Liquor/h, Weight \\
\textbf{Treatment variables}  &  packed red blood cells, FFP, platelets, coagulation factors, norepinephrine, epinephrine, dobutamine, milrinone, levosimendan, theophyllin, vasopressin, desmopressin, vasodilatators, ACE Inhibitors, Sartane, Ca Antagonists, B-Blocker, Andere, Adenosin, Digoxin, Amiodaron, Atropin, K-sparend, Aldosteron Antagonist, Loop diuretics, Thiazide, Acetazolamide, Administriation of antibiotics, Administation of antimycotic, administration of antiviral, Antihelmenticum, Benzodiacepine, Alpha 2 Agonisten, Barbiturate, Propofol, Glucose Administration, Insuling Langwirksam, Insulin Kurzwirksam, Nimodipin, Opiate, Non-opioide, NSAR, Ketalar, Peripherial Anesthesia, Steroids, Thrombozytenhemmer, Enteral Feeding, Parenteral Feeding, Heparin, NMH, Others in Case of HIT, Marcoumar, Protamin, Anti Fibrinolyticum, Kalium, Phosphat, Na, Mg, Ca, Trace elements, Bicarbonate, Antiepileptica, Anti delirant medi, Psychopharma, steroids, non-steroids, Thyrotardin, Thyroxin, Thyreostatikum, Mineralokortikoid, Antihistaminka, Chemotherapie, Lysetherapie, Muskelrelaxans, Anexate, Naloxon, Beh. Pulm. Hypertonie, Pankreas Enzyme, Terlipressin, Immunoglobulin, Immunsuppression, VitB Substitution, Parkinson Medikaiton, GCSF \\
\bottomrule
    \end{tabular}}
\end{table*}
 \begin{table*}[t]
    \caption{\textbf{Variable splitting by data acquisition type,} obtained based on public metadata in the MIMIC-III dataset.}
    \label{tab:type_splitting_m}
    \resizebox{\textwidth}{!}{\begin{tabular}{l p{12cm}}
    \toprule
        Variable type & Variable name \\ \midrule
        \textbf{Laboratory values}  &  Glucose, pH \\
        \textbf{Monitored variables}  &  Diastolic blood pressure, Heart Rate, Mean blood pressure, Systolic blood pressure, Temperature, Fraction inspired oxygen, Oxygen saturation, Respiratory rate \\
        \textbf{Observed variables}  &  Glascow coma scale eye opening, Glascow coma scale motor response, Glascow coma scale total, Glascow coma scale verbal response, Capillary refill rate \\
    \bottomrule
    \end{tabular}}
\end{table*} \newpage
\begin{table*}[htbp]
\floatconts
{tab:ablation_resp}
{\caption{\textbf{Benchmarking analysis of embedding design choices} for \emph{Respiratory failure prediction} on the HiRID dataset. Ablations on the default architecture: FTT \citep{gorishniy2021revisiting} with organ splitting and attention-based aggregation.}\vspace{-0.5cm}}  
    {
\subtable[Embedding architecture.][b]{
\begin{tabular}{lc}
        \toprule
        Architecture & AUPRC \\
        \midrule
             None &  59.5 $\pm$ 0.4 \\
MLP & \textbf{60.6} $\pm$ 0.2\\
             ResNet & 58.2 $\pm$ 0.4\\
             FTT  & \textbf{60.7} $\pm$ 0.5\\
         \bottomrule
    \end{tabular}}\qquad
\subtable[Group aggregation.][b]{\begin{tabular}{lc}
        \toprule
        Aggregation & AUPRC \\
        \midrule
             Concatenate & \textbf{61.1} $\pm$ 0.1 \\
             Average & 60.1 $\pm$ 0.3 \\
             Attention & \textbf{60.7} $\pm$ 0.2\\
         \bottomrule
    \end{tabular}}\qquad
\subtable[Feature grouping.][b]{\begin{tabular}{lc}
        \toprule
        Grouping & AUPRC \\
        \midrule
        None & 59.8 $\pm$ 0.1\\
             Variable type &  \textbf{60.7} $\pm$ 0.1\\
Meas. type & \textbf{60.3} $\pm$ 0.3\\
Organ & \textbf{60.7} $\pm$ 0.5\\
         \bottomrule
    \end{tabular}}
}
\end{table*} \begin{table*}[htbp]
\floatconts
{tab:ablation_mort}
    {\caption{\textbf{Benchmarking analysis of embedding design choices} for \emph{Mortality prediction} on the HiRID dataset. Ablations on the default architecture: FTT \citep{gorishniy2021revisiting} with organ splitting and attention-based aggregation.}\vspace{-0.5cm}} 
    {
    \subtable[Embedding architecture.][b]{
    \begin{tabular}{lc}
        \toprule
        Architecture & AUPRC \\
        \midrule
             None &  60.1 $\pm$ 0.3 \\
MLP & {60.3} $\pm$ 0.2\\
             ResNet & 57.8 $\pm$ 0.4\\
             FTT  & \textbf{61.6} $\pm$ 1.3\\
         \bottomrule
    \end{tabular}}\qquad
    \subtable[Group aggregation.][b]{\begin{tabular}{lc}
        \toprule
        Aggregation & AUPRC \\
        \midrule
             Concatenate & \textbf{62.3} $\pm$ 1.9 \\
             Average & 61.0 $\pm$ 0.7 \\
             Attention & 61.6 $\pm$ 1.3\\
         \bottomrule
    \end{tabular}}\qquad
    \subtable[Feature grouping.][b]{\begin{tabular}{lc}
        \toprule
        Grouping & AUPRC \\
        \midrule
        None & 60.5 $\pm$ 0.6\\
             Variable type &  {60.9} $\pm$ 0.2\\
Meas. type & 61.6 $\pm$ 1.0\\
             Learned &  \textbf{62.3} $\pm$ 1.2 \\
             Organ & \textbf{61.6} $\pm$ 1.3\\
         \bottomrule
    \end{tabular}}
    
    }
\end{table*} \begin{table*}[htbp]
\floatconts
 {tab:ablation_ftt_los_hirid}
 {\caption{\textbf{Benchmarking analysis of embedding design choices} for \emph{Length-of-Stay} prediction on the HiRID dataset. Best performing model shown while fixing the specific variation and performing a random search over the others.}\vspace{-0.5cm}}
 {\subtable[Embedding architecture.][b]{\begin{tabular}{lc}
        \toprule
        Architecture & MAE $\downarrow$ \\
        \midrule
             None &  59.3 $\pm$ 0.6 \\
             MLP  & 56.9 $\pm$ 1.1 \\ ResNet & 57.3 $\pm$ 0.7 \\ FTT & \textbf{54.0} $\pm$ 0.1 \\ \bottomrule
    \end{tabular}
   }\qquad
   \subtable[Group aggregation.][b]{\begin{tabular}{lc}
        \toprule
        Aggregation & MAE $\downarrow$ \\
        \midrule
             Concatenate & 54.2 $\pm$ 0.2 \\
             Average & \textbf{54.0 $\pm$ 0.1} \\
             Attention & \textbf{54.0 $\pm$ 0.1} \\
         \bottomrule
    \end{tabular}
   }\qquad
   \subtable[Feature grouping][b]{\begin{tabular}{lc}
        \toprule
        Grouping & MAE $\downarrow$ \\
        \midrule
            None & 55.7 $\pm$ 0.1 \\
Meas. type & 54.4 $\pm$ 0.3 \\
Organ & \textbf{54.0} $\pm$ 0.1 \\
         \bottomrule
    \end{tabular}
   }
 }
\end{table*} \begin{table*}[htbp]
\floatconts
 {tab:ablation_ftt_pheno_hirid}
 {\caption{\textbf{Benchmarking analysis of embedding design choices} for \emph{Phenotyping} prediction on the HiRID dataset. Best performing model shown while fixing the specific variation and performing a random search over the others.}\vspace{-0.5cm}}
 {\subtable[Embedding architecture.][b]{\begin{tabular}{lc}
        \toprule
        Architecture & \textit{Bal.Acc} $\uparrow$ \\
        \midrule
             None &  42.7 $\pm$ 1.5 \\
             MLP  & 39.5 $\pm$ 1.8 \\ ResNet & 43.3 $\pm$ 1.7 \\ FTT & \textbf{46.5} $\pm$ 1.4 \\ \bottomrule
    \end{tabular}
   }\qquad
   \subtable[Group aggregation.][b]{\begin{tabular}{lc}
        \toprule
        Aggregation & \textit{Bal.Acc} $\uparrow$ \\
        \midrule
             Concatenate & 43.2 $\pm$ 0.9 \\
             Sum & \textbf{46.5} $\pm$ 1.4 \\
             Attention & 41.8 $\pm$ 1.7 \\
         \bottomrule
    \end{tabular}
   }\qquad
   \subtable[Feature grouping][b]{\begin{tabular}{lc}
        \toprule
        Grouping & \textit{Bal.Acc} $\uparrow$ \\
        \midrule
            None & 39.8 $\pm$ 2.6 \\
Meas. type & 43.6 $\pm$ 0.8 \\
Organ & \textbf{46.5} $\pm$ 1.4 \\
         \bottomrule
    \end{tabular}
   }
 }
\end{table*} 
\begin{table*}[htbp]
\floatconts
{tab:ablation_decomp}
    {\caption{\textbf{Benchmarking analysis of embedding design choices} for \emph{Decompensation prediction} on MIMIC-III dataset. Ablations on the default architecture: FTT \citep{gorishniy2021revisiting} with organ splitting and attention-based aggregation.}\vspace{-0.5cm}}
    {
    \subtable[Embedding architecture.][b]{
    \begin{tabular}{lc}
        \toprule
        Architecture & AUPRC \\
        \midrule
             None &  \textbf{38.7} $\pm$ 0.3 \\
MLP & {36.3} $\pm$ 0.3\\
FTT  & 38.0 $\pm$ 0.4\\
         \bottomrule
    \end{tabular}}\qquad    
    \subtable[Group aggregation.][b]{\begin{tabular}{lc}
        \toprule
        Aggregation & AUPRC \\
        \midrule
             Concatenate &  36.2 $\pm$ 1.3 \\
             Average &  37.4 $\pm$ 0.1 \\
             Attention &  \textbf{38.0} $\pm$ 0.4 \\
         \bottomrule
    \end{tabular}}\qquad
    \subtable[Feature grouping.][b]{\begin{tabular}{lc}
        \toprule
        Grouping & AUPRC \\
        \midrule
        None & \textbf{38.7} $\pm$ 0.3\\
             Variable type &  {34.8} $\pm$ 0.3\\
Meas. type & 38.1 $\pm$ 0.2\\
             Organ & 38.0 $\pm$ 0.4\\
         \bottomrule
    \end{tabular}}
    \caption{Feature grouping.} }
\end{table*} \begin{table*}[htbp]
\floatconts
{tab:ablation_ihm}
    {\caption{\textbf{Benchmarking analysis of embedding design choices} for \emph{Mortality prediction} on MIMIC-III dataset. Ablations on the default architecture: FTT \citep{gorishniy2021revisiting} with organ splitting and attention-based aggregation.} \vspace{-0.5cm}}   
    {
    \subtable[Embedding architecture.][b]{
    \begin{tabular}{lc}
        \toprule
        Architecture & AUPRC \\
        \midrule
             None &  51.2 $\pm$ 0.8 \\
MLP & 51.3 $\pm$ 1.01 \\
             ResNet & 50.6 $\pm$ 0.7\\
             FTT  & \textbf{51.8} $\pm$ 0.6\\
         \bottomrule
    \end{tabular}}\qquad
    \subtable[Group aggregation.][b]{\begin{tabular}{lc}
        \toprule
        Aggregation & AUPRC \\
        \midrule
             Concatenate &  51.9 $\pm$ 0.6\\
             Average &  \textbf{52.6} $\pm$ 0.6\\
             Attention & 51.8 $\pm$ 0.6\\
         \bottomrule
    \end{tabular}}\qquad
    \subtable[Group aggregation.][b]{
    \begin{tabular}{lc}
        \toprule
        Grouping & AUPRC \\
        \midrule
        None & 51.1 $\pm$ 0.5\\
             Variable type &  51.1 $\pm$ 0.7\\
Meas. type & 51.4 $\pm$ 2.2 \\
Organ & \textbf{51.8} $\pm$ 0.6\\
         \bottomrule
    \end{tabular}}
    }
\end{table*}

\end{document}